\documentclass[]{interact}

\usepackage{epstopdf}% To incorporate .eps illustrations using PDFLaTeX, etc.
\usepackage[caption=false]{subfig}% Support for small, `sub' figures and tables
\usepackage[numbers, compress]{natbib}% Citation support using natbib.sty
\bibpunct[, ]{[}{]}{,}{n}{,}{,}% Citation support using natbib.sty
\usepackage{array}
\usepackage{float}
\usepackage[utf8]{inputenc}
\usepackage{multirow}
\usepackage{url}
\usepackage{float}
\usepackage{xcolor}
\usepackage[normalem]{ulem}
\usepackage{soul}

\theoremstyle{plain}% Theorem-like structures provided by amsthm.sty

\theoremstyle{definition}

\theoremstyle{remark}

\begin{document}

\title{Toward AI-Friendly Cartography: Understanding How Color Design Influences Foundation Model Spatial Reasoning on Sequential Choropleth Maps}

\author{ 
  Yonghe Sun\textsuperscript{a,b}, 
  Zhenjia Liu\textsuperscript{a,b}, 
  Hua Liao\textsuperscript{a,b},
  Wenjia Xu\textsuperscript{c}, 
  Nai Yang\textsuperscript{d}, 
  Weihua Dong\textsuperscript{e},  
  Zhiwei Wei\textsuperscript{a,b*} \thanks{CONTACT Zhiwei Wei. Email: trentonwei@whu.edu.cn. ORCID: https://orcid.org/0000-0002-3494-3686}
  \\
  \affil{\textsuperscript{a}School of Geographic Sciences, Hunan Normal University, Changsha, China.}
  \affil{\textsuperscript{b}Hunan Key Laboratory of Geospatial Big Data Mining and Application, Changsha, China.}
  \affil{\textsuperscript{c}School of Information and Communication Engineering, Beijing University of Posts and Telecommunications, Beijing, China.}
  \affil{\textsuperscript{d}School of Geography and Information Engineering, China University of Geosciences, Wuhan, China.}
  \affil{\textsuperscript{e}Advanced Interdisciplinary Institute of Satellite Applications, State Key Laboratory of Earth Surface Processes and Resource Ecology, Faculty of Geographical Science, Beijing Normal University, Beijing, China.}
}

\maketitle

\begin{abstract}
Recent advances in foundation models (FMs) have significantly improved multimodal reasoning and geospatial understanding, leading to growing interest in map-based spatial cognition for FMs. However, most existing cartographic design principles were originally developed for human visual perception, and it remains unclear whether these principles are equally effective for FM reasoning. To address this gap, we focus on choropleth maps, one of the most widely used forms of thematic cartography, and systematically investigate how three classical cartographic factors--sequential hue palettes, sequential versus randomized color ordering, and lightness contrast--influence FM spatial reasoning. We construct a large-scale benchmark containing 5,760 choropleth maps with controlled spatial structures, together with 28,800 multi-level spatial reasoning tasks spanning \emph{Attribute Identify}, \emph{Spatial Recognition}, \emph{Compare}, \emph{Rank}, and \emph{Pattern Delineate}. Twenty-one recent multi-modal FMs from both open-source and proprietary ecosystems are systematically evaluated. Experimental results reveal: First, different sequential hue palettes produce only a limited and non-systematic influence on model performance, suggesting that FMs rely less on hue semantics than human map readers. Second, contrary to conventional assumptions, disrupting sequential color ordering substantially degrades spatial reasoning performance across most evaluated models, particularly for comparison- and ranking-based tasks. Third, lightness contrast constitutes a fundamental machine-readable signal: reducing contrast consistently harms reasoning performance, whereas further increasing contrast provides only marginal additional improvement once sufficient separability is achieved. To further assess whether these sensitivities are intrinsic or can be learned, we additionally perform lightweight LoRA fine-tuning, which substantially improves overall geospatial reasoning while confirming that the relative sensitivity to sequential ordering and lightness contrast remains stable. These findings indicate that FMs depend heavily on conventional sequential color ordering and sufficient contrast during map understanding. Overall, our results suggest that some classical cartographic principles remain highly beneficial for machine spatial reasoning, while others transfer less effectively to machine cognition, providing empirical guidance for AI-friendly cartography.
\end{abstract}


\section{Introduction}

Recent developments in large-scale foundation models (FM) have demonstrated extraordinary capabilities in language understanding, multi‑modal reasoning, and problem solving across diverse domains, with models such as GPT‑4 achieving human‑level performance on professional and academic benchmarks (e.g., passing a simulated bar exam and exhibiting advanced multi‑domain reasoning) \cite{achiam2023gpt, schulze2025visual}. These capabilities have also attracted the attention of geographers and spatial scientists, giving rise to the field of GeoAI, where FMs are increasingly applied to geographic tasks such as urban planning analysis, spatial pattern recognition, environmental monitoring, and policy evaluation \cite{janowicz2025geofm, yang2025evaluating, zheng2025urban}. Within these tasks, maps play a central role by representing spatial structure, visualizing regional attributes, and revealing patterns and trends \cite{wang2026map}. Consequently, enabling FMs to accurately interpret maps has become a critical challenge for the development of AI systems capable of robust geospatial reasoning.

Building on this growing interest, recent work has applied FMs to both map understanding and map-guided applications. At the foundational level, work such as MapLayNet \cite{yang2025maplaynet}, layout analyses in academic cartography \cite{wei2026evolving}, and raster text recognition focus on extracting structural and semantic information from maps \cite{chiang2015recognizing}; While MapReader \cite{zhang2025mapreader} and related datasets like MapQA \cite{li2025benchmarking}, CartoMark \cite{zhou2024cartomark}, and MapVerse \cite{bhat2026mapverse} enable question-answering (QA) over map content. Complementing these, benchmarks such as GVSABench \cite{liu2026can} and GeoAnalystBench \cite{zhang2025geoanalystbench} assess multimodal models’ ability to reason and analyze maps. In parallel, research on AI-assisted cartography, including CartoAgent \cite{wang2025cartoagent}, MapColorAI \cite{yang2025mapcolorai}, MapGPT \cite{zhang2024mapgpt}, and transformer-based contour map representation learning \cite{kong2026tokenization}, explores how FMs can support design, automated generation, and color encoding of maps. Collectively, these studies demonstrate the potential of FMs to read, reason about, and interact with maps.

However, these advances primarily focus on task performance or application; it remains unclear how fundamental map design principles influence machine (FM) reasoning. Most existing map design conventions, such as hue variation, color sequence, and contrast in choropleth maps, were developed to optimize human visual perception \cite{Brewer02Pic, Brewer03, wu2024computational}. However, the cognitive mechanisms of FMs differ substantially from humans: for example, visual tokens are processed patch-wise, and attention mechanisms may respond differently to gradients and contrasts than the human eye \cite{wang2025emulating}. Consequently, principles that are effective for humans may not translate directly to machine understanding. Systematically evaluating these principles is therefore crucial: understanding which design elements constitute core signals for machines and which are auxiliary can provide concrete guidance for AI‑friendly map design, enabling models to more accurately interpret spatial structures, regional attributes, and patterns. This suggests that improving machine spatial reasoning may not solely depend on modifying models themselves, but can also be achieved by redesigning the visual representations that models interpret. Motivated by this perspective, our study investigates which map design principles are effectively AI-friendly and how they influence machine spatial reasoning.

To address the above question, we focus on choropleth maps, which are widely used to represent quantitative attributes across geographic regions, such as the U.S. election results \cite{maceachren2004maps, dent1995cartography}.  Among the various design elements in choropleth maps, color serves as the most salient visual channel for conveying quantitative information. Over decades, cartographers have developed standardized design practices for these maps, such as the famous ColorBrewer framework, which provides Sequential, Diverging, and Qualitative color templates \cite{Brewer02Pic, Brewer03, wu2022adaptive}. Building on these conventions, our study focuses on three key aspects of sequential choropleth map color design that may influence machine reasoning: color hue, sequential ordering, and color difference magnitude. To investigate these factors, we constructed controlled sets of 5760 choropleth maps based on ColorBrewer sequential templates, systematically varying one factor at a time. The benchmark encompasses multiple cognitive dimensions, including \emph{Attribute Identify}, \emph{Spatial Recognition}, \emph{Compare}, \emph{Rank}, and \emph{Pattern Delineate}, covering tasks from fine-grained local perception to holistic spatial reasoning. 21 models are evaluated on these tasks using large vision-language models, either open-source or proprietary. Performance is analyzed using multi-factor statistical methods to determine which aspects of map design—color hue, sequential ordering, and color difference magnitude—serve as core signals for machines and which are secondary, ultimately informing AI‑friendly map design practices. Our main contributions are threefold:

\begin{itemize}

    \item We present one of the first systematic investigations of how classical choropleth color design principles influence FM spatial reasoning, bridging traditional cartography and the emerging direction of AI-friendly cartography.

    \item We construct a large-scale controlled benchmark containing 5,760 choropleth maps and 28,800 spatial reasoning tasks, systematically covering variations in hue palettes, sequential ordering, and lightness contrast across five cognitive dimensions: \emph{Attribute Identify}, \emph{Spatial Recognition}, \emph{Compare}, \emph{Rank}, and \emph{Pattern Delineate}.

    \item We reveal that different cartographic color principles contribute unequally to machine spatial reasoning. In particular, sequential ordering and lightness contrast substantially influence FM performance, whereas hue variation produces only limited effects. Based on these findings, we provide empirical insights into machine-oriented thematic map design and AI-friendly cartography.

\end{itemize}

\section{Related work}
\label{relatedwork}

\subsection{Choropleth Map Design and Applications}

Choropleth maps are a widely used form of thematic cartography that visually encode quantitative data across spatial regions. The earliest choropleth maps can be traced back to the early 19th century, with Charles Dupin's 1826 map of literacy rates in France widely recognized as one of the first systematic choropleth maps \cite{palsky2008connections}. This method then quickly gained popularity as census data became more widely available in Europe and North America. Early cartographic work emphasized perceptual clarity and effective representation of spatial data, introducing the foundation for systematic consideration of visual variables such as color \cite{maceachren2004maps, dent1995cartography}. These studies established the early principles of thematic map design and highlighted the need to communicate quantitative differences across regions clearly.

Building on these foundations, formal design research emerged in the latter half of the 20th century, focusing on systematic guidelines for color use and perception. Brewer’s pioneering work on color charts highlighted practical challenges of palette selection and emphasized perceptually grounded sequences for both sequential and diverging data representations \cite{Brewer03, Brewer94}. Subsequent studies, including Brewer et al.’s evaluation of mortality mapping color schemes, combined theoretical review with empirical testing to determine which color combinations support accurate map reading and user preferences \cite{Brewer02Pic}. Comprehensive cartographic overviews further detailed the trade-offs among classification methods, color schemes, and visual clarity, while research on complex designs, such as bivariate choropleth maps, emphasized the need to align color choices with spatial relationships and analytical questions \cite{Slocum2022}. Complementing these design-oriented studies, empirical investigations into human perception have provided quantitative validation of these principles. For example, controlled experiments \cite{Schiewe19} demonstrated that users rely heavily on color lightness to detect extreme values and spatial patterns, and that biases such as dark‑is‑more, area‑size, and data-classification effects significantly influence interpretation accuracy. Follow‑up work has shown that these perceptual biases persist under different visual conditions, such as dark mode displays \cite{Schiewe24}. Cognitive research also suggests that humans develop expectations (e.g., darker colors indicate larger magnitudes) when interpreting colormap data visualizations, and deviations from these expectations can lead to misinterpretation \cite{Soto23}. Quantitative experiments have further investigated appropriate hue ranges for sequential color schemes, demonstrating that certain hue intervals improve performance on identification, comparison, and ranking tasks \cite{Chen25}. Additional perceptual work indicates that legend design and range settings influence users’ judgments of absolute magnitudes \cite{Bradley24}. Together, these studies provide a coherent empirical foundation for evaluating the effectiveness of choropleth map color schemes, linking design principles with human cognitive responses.

To support practical adoption of these design principles, standardized frameworks and tools have also been developed. Early software applications provided basic support for selecting color schemes and classification methods, allowing mapmakers to apply consistent palettes and reduce perceptual errors, such as the famous ColorBrewer tool \cite{harrower2003colorbrewer}. Alongside such general tools, cartographic research has proposed methods to systematically enhance map legibility through improved color contrast and knowledge‑based specification. For example, Chesneau (2011) developed a model for the automatic improvement of colour contrasts in maps, specifically applied to risk maps; the model iteratively identifies poorly contrasted elements and adjusts them using a schema of colour contrast rules, demonstrating measurable gains in map readability \cite{chesneau2011model}. Similarly, Christophe (2011) proposed the COLorLEGend system, which integrates knowledge from visual perception, semiotics, and cartographic rules to help users create personalized and harmonious color specifications \cite{christophe2011creative}. As computing capabilities have advanced, more sophisticated frameworks such as GeoLinter and GeoExplainer have emerged, enabling systematic evaluation and improved interpretability of map designs against established perceptual and cartographic rules \cite{lei2023geolinter, lei2023geoexplainer}. Wu et al. (2024) also proposed a computational framework for assessing the aesthetic quality of map colors by integrating cartographic aesthetic principles, spatial organization, and computational aesthetic metrics, demonstrating that map color aesthetics can be quantitatively modeled and predicted with high accuracy \cite{wu2024computational}. In parallel, AI‑assisted approaches like MapColorAI and MapColor-Agent leverage the large language model to adaptively generate color schemes tailored to specific data distributions and visualization goals, combining statistical analysis with perceptual optimization \cite{yang2025mapcolorai, wei2026mapcoloragent}. Other tools, such as automated contour map representation using transformer‑based encoders, focus on representing spatial boundaries and gradient information in a machine‑readable format, further bridging design principles with scalable map production and analysis \cite{kong2026tokenization}. Complementing these design-oriented and generative approaches, a parallel line of work applies machine learning directly to extracting map content, particularly where color plays a central symbolic role. DIGMAPPER provides a modular system for automated digitization of geologic maps, integrating multiple deep learning components for symbol and boundary extraction \cite{duan2025digmapper}. More specifically targeting \textbf{color-oriented} content extraction, Luo et al. (2023) apply deep learning to extract critical-minerals features from geological maps based on color-coded symbology \cite{luo2023critical}, and Lin et al. (2023) exploit polygon metadata to accurately extract polygonal features from raster maps, addressing challenges posed by color and pattern variation in map legends \cite{lin2023exploiting}. These extraction-focused studies further underscore that color remains a fundamental, if often implicit, channel for map content interpretation, reinforcing the motivation for systematically examining color design principles in the context of machine reasoning.

In summary, existing research has established a solid foundation for choropleth map design, combining empirical studies, perceptual principles, and standardized frameworks. These developments illustrate a clear progression from manual, expert-driven design to semi-automated and AI-assisted workflows. However, the majority of these efforts remain focused on human-centered design, and systematic investigation into machine-oriented or AI-friendly map design principles--especially regarding how maps can be optimized for automated interpretation and spatial reasoning--still remains largely unexplored.

\subsection{Large Models for Map Understanding and Reasoning}

With the rapid development of FMs, they have gradually been applied to the domain of maps, supporting a variety of spatial reasoning and geospatial analysis tasks. Research in this area can be broadly categorized into two streams: (1) map content understanding, and (2) map-based applications.

The first line of research focuses on \textbf{map content understanding}, such as map element recognition and map-based question answering. Regarding map element recognition, models such as MapLayNet have been developed to extract layout information, including spatial arrangements and hierarchies of map elements, enabling systematic parsing of map structures \cite{yang2025maplaynet}. Similarly, Wei et al. (2026) analyzed layout trends across multi-lingual journal maps over time by automatically extracting map elements via the YOLO series and SAM models, revealing recurring structural patterns and design conventions \cite{wei2026evolving}. For textual information embedded in raster maps, Chiang and Knoblock (2015) introduced a semi-automatic method using example text areas and cartographic labeling principles to locate and rotate labels before OCR, improving text recognition in heterogeneous raster maps \cite{chiang2015recognizing}. Ma et al. (2023) proposed a CNN-based method for automatic extraction of depth annotations in charts \cite{mengkai2023automatic}. With the advancement of FMs, these techniques were extended to large vision-language models; for instance, \emph{MapReader} demonstrates the use of such models for answering geospatial queries directly from map images \cite{zhang2025mapreader}, and Xu and Tao (2024) evaluated GPT-4V's ability on map reading and analysis tasks \cite{xu2024map}. Beyond general-purpose evaluation, MapBench introduces a navigation-oriented benchmark of over 1,600 path-finding problems across 100 diverse maps, requiring LVLMs to generate language-based navigation instructions and revealing substantial gaps relative to human wayfinding performance \cite{xing2025can}. Complementing these general map-reading evaluations, language models have also been applied to historical map understanding: Liu et al. (2025) developed an automatic map storytelling system for historical maps \cite{liu2025efficient}, while related work integrates spatio-temporal knowledge graphs with large language models to support geospatial question answering over historical map collections \cite{liu2025geospatial}. To support system evaluation, datasets such as MapQA, GVSABench, and CartoMark provide curated pairs of map images and questions for training and evaluation \cite{li2025benchmarking, zhou2024cartomark, liu2026can}, while benchmarks like GeoAnalystBench evaluate multimodal models across a range of spatial tasks \cite{zhang2025geoanalystbench}. A closely related benchmark is FRIEDA~\cite{pyo2025frieda}, which evaluates multi-step cartographic reasoning in LVLMs using heterogeneous real-world maps from domains such as geology, urban planning, and environmental studies. FRIEDA requires models to interpret legends, map text, topology, direction, and distance, and demonstrates that current state-of-the-art LVLMs still perform substantially below humans on complex real-world map reasoning. Although these studies show that models can interpret map content and perform spatial reasoning, they primarily focus on task performance rather than the impact of underlying map design principles.

The second line explores \textbf{map-based applications leveraging FMs}. For example, \emph{CartoAgent} provides AI-assisted tools for optimizing map symbol design \cite{wang2025cartoagent}, whereas \emph{MapColorAI} uses large language models to adaptively generate color schemes and improve visual clarity for choropleth maps \cite{yang2025mapcolorai}. Focusing on administrative maps, Wei et al. (2026) integrate large language models with a multi-agent collaboration mechanism to perform task decomposition and user-guided color scheme generation \cite{wei2026mapcoloragent}. For automatic map generation, \emph{MapGPT} enables creation of maps from structured geospatial data, facilitating end-to-end production \cite{zhang2024mapgpt}. Affolter et al. (2025) developed a generative AI framework that integrates vector data with diffusion-based image generation models to produce maps in controlled styles from textual prompts, enabling both experts and non-experts to efficiently create accurate and customizable maps \cite{affolter2025generative}. Additionally, transformer-based contour map representations encode spatial boundaries and gradient information in a machine-readable format, supporting downstream reasoning and analysis \cite{kong2026tokenization}. These approaches illustrate how AI-assisted systems can enhance both the production and analytical interpretation of maps.

Collectively, these studies provide a strong foundation for understanding how FMs process maps. However, they have primarily focused on task accuracy and automated generation; a systematic evaluation of \textbf{which map design principles, originally developed for human cognition, are effective for machine reasoning} still remains largely unexplored. Addressing this gap is essential for establishing AI-friendly map design guidelines that optimize map readability for both human users and machine reasoning, and it constitutes the primary motivation for this work.

\section{Methodology}
\label{method}

\subsection{Overview of the Framework}

\begin{figure}[H]
	\centering
	\makebox[\textwidth][c]{
        \includegraphics[width=1.2\textwidth]{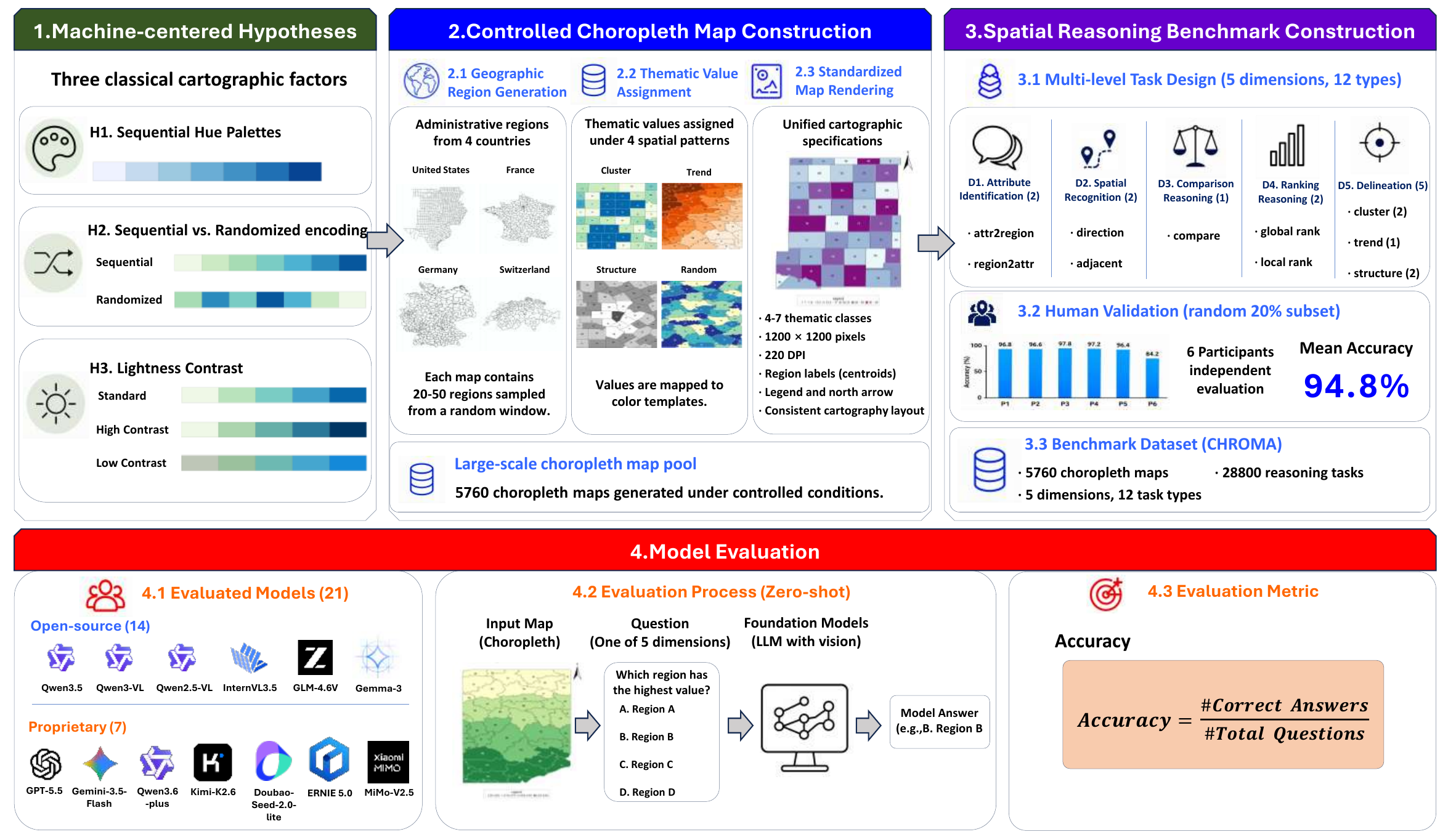}
    }
        \captionsetup{skip=0pt}  % 调整该图的上下间距
	\caption{The overview of the framework.}
	\label{framework}
\end{figure}

The overall framework of this study is designed to systematically investigate how classical cartographic color design factors influence FM spatial reasoning. The framework consists of four main stages: machine-centered hypotheses (Sec.~\ref{hypothesis}), controlled choropleth map generation (Sec.~\ref{mapCons}), benchmark construction (Sec.~\ref{benchmark}), and model evaluation (Sec.~\ref{evaluation}), as shown in Figure~\ref{framework}. Specifically, the machine-centered hypotheses stage formulates three research hypotheses focusing on sequential hue palettes (H1), sequential versus randomized color ordering (H2), and lightness contrast (H3). The controlled choropleth map generation stage designs corresponding cartographic color templates based on H1-H3 and constructs large-scale choropleth maps with the designed visual manipulations. The benchmark construction stage designs multi-level spatial reasoning tasks and conducts human validation to ensure benchmark reliability and interpretability. Finally, the model evaluation stage systematically evaluates 21 multimodal FMs under unified experimental settings to analyze how different cartographic color designs influence machine spatial reasoning performance.

\subsection{Machine-centered Hypotheses}
\label{hypothesis}

Based on the summary of choropleth maps and the visual processing mechanisms of large vision-language models in Sec.~\ref{relatedwork}, we formulate three hypotheses regarding how sequential choropleth color encoding influences machine spatial reasoning. Specifically, we investigate the effects of \textbf{color hue} (H1), \textbf{sequential ordering} (H2), and \textbf{color contrast} (H3), as follows.

\textbf{H1: Different hue palettes will have limited influence on FM spatial reasoning performance.}

In traditional cartography, different sequential hue palettes are widely used to improve aesthetics, thematic distinction, and perceptual harmony \cite{Brewer03, Brewer94}. However, for FMs, map images are ultimately represented as visual tokens and pixel-level patterns rather than semantic color concepts \cite{wang2025emulating}. As a result, variations in hue may provide limited additional information for machine spatial reasoning.

\textbf{H2: Sequential color ordering may exert a weaker influence than expected from human cartographic perception.}

Sequential ordering is one of the most fundamental principles in sequential choropleth map design because humans naturally associate monotonic luminance progression with ordered quantitative magnitude \cite{Brewer03, Slocum2022}. However, unlike humans, FMs do not explicitly perceive cartographic semantics or ordinal symbolism. Instead, they primarily process local visual patterns through patch-based representations and attention mechanisms \cite{dreyer2025mechanistic}. As a result, disrupting sequential ordering may not substantially reduce model performance if sufficient local visual separability is preserved.

\textbf{H3: Increasing color contrast between adjacent thematic classes improves FM spatial reasoning performance.}

Previous cartographic and perceptual studies have shown that color contrast, especially luminance contrast, strongly affects human ability to distinguish thematic classes and identify spatial patterns \cite{Schiewe19, Schiewe24}. Unlike humans, FMs primarily encode map images into patch-level visual tokens and rely on differences between local visual features during representation learning \cite{dreyer2025mechanistic}. Stronger color contrast may enlarge feature differences between neighboring regions, making thematic boundaries and spatial transitions more distinguishable during visual encoding and attention aggregation. Therefore, compared with hue and sequential semantics, color contrast may constitute a more fundamental signal for machine spatial reasoning.

\subsection{Controlled Choropleth Map Dataset Construction}
\label{mapCons}

To systematically evaluate the three machine-centered hypotheses in Sec.~\ref{hypothesis}, we constructed a controlled pool of choropleth maps using hypothesis-specific color templates, with corresponding designs detailed in Secs.~\ref{H1-generated}, \ref{H2-generated}, and \ref{H3-generated}. All maps were generated through a unified and fully controlled pipeline, sharing identical geographic regions, spatial structures, thematic distributions, rendering settings, and layout configurations unless explicitly manipulated by the target hypothesis. Details of region generation, thematic assignment, and map rendering are provided in Sec.~\ref{Map-generated}.

\subsubsection{H1: Hue-controlled}
\label{H1-generated}

This experiment evaluates whether variations in hue palettes influence FM spatial reasoning performance (H1). To this end, we construct a hue-controlled benchmark based on sequential color templates. Sequential palettes have been extensively studied and standardized in cartography, particularly through the ColorBrewer framework \cite{Brewer03}. Following its original categorization, we select all 18 representative sequential palettes, including both single-hue and multi-hue schemes, as summarized in Table~\ref{tab:hue_palettes}. To avoid overly simplistic maps with limited thematic distinction and overly complex maps with excessive visual categories, the number of classes is restricted to 4--7, consistent with commonly recommended practices in choropleth cartography \cite{Brewer03, Slocum2022, zhiwei2018backtracking}.

\begin{table}[H]
\centering
\scriptsize
\caption{Sequential color palettes selected from ColorBrewer for the hue-controlled benchmark.}
\label{tab:hue_palettes}
\makebox[\linewidth][c]{
\begin{tabular}{ll}
\toprule
Category & Sequential palettes \\
\midrule
Single-hue &
Blues, Greens, Greys, Oranges, Purples, Reds \\
\midrule
Multi-hue &
BuGn, BuPu, GnBu, OrRd, PuBu, PuBuGn,
PuRd, RdPu, YlGn, YlGnBu, YlOrBr, YlOrRd \\
\bottomrule
\end{tabular}
}
\end{table}

\subsubsection{H2: Sequential vs Randomized Encoding}
\label{H2-generated}

\begin{figure} [H]
	\centering
	\includegraphics[width=\textwidth]{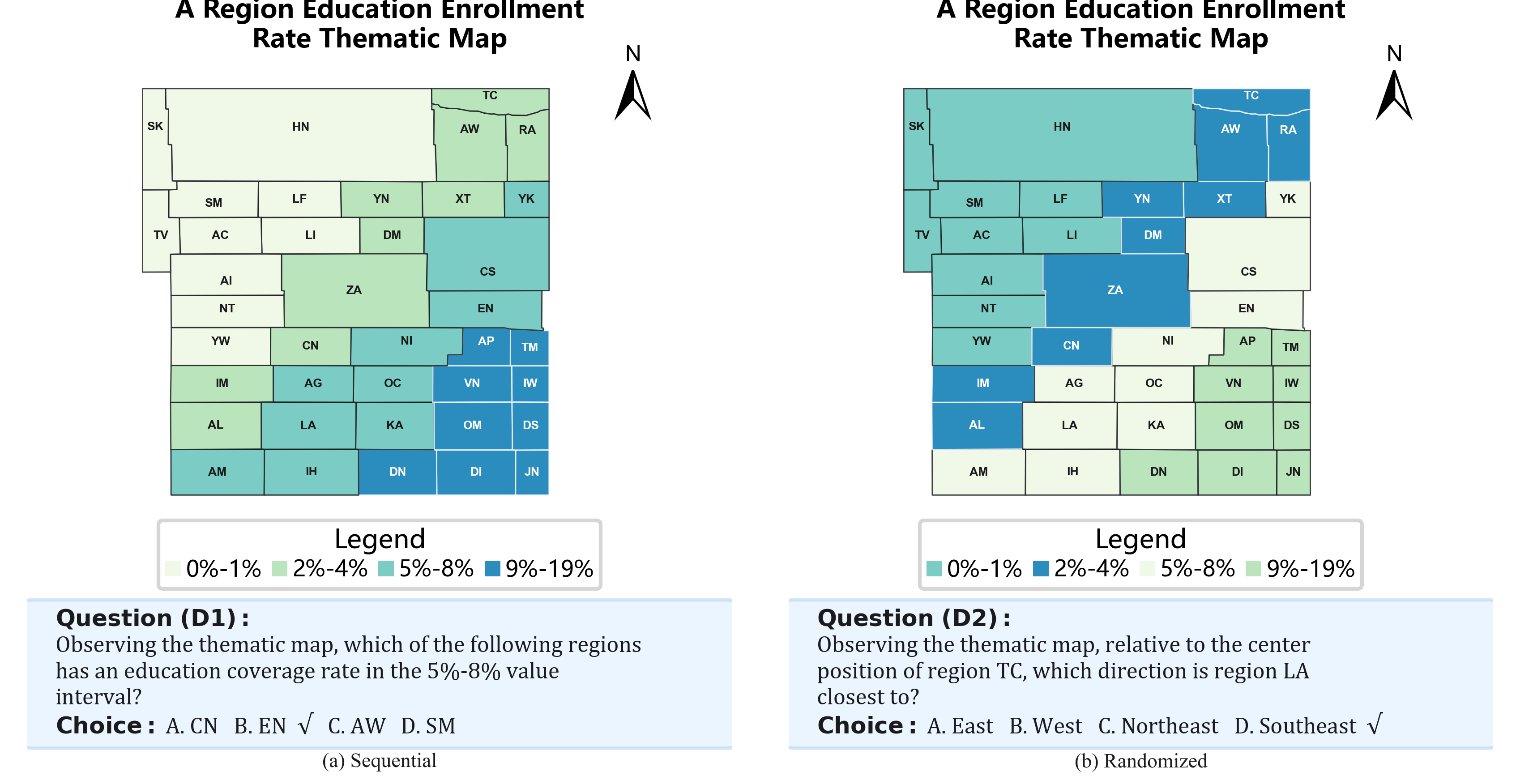}
        \captionsetup{skip=0pt}  % 调整该图的上下间距
	\caption{Sequential(a) and randomized(b) choropleth encoding under identical spatial distributions. The randomized condition preserves the original color set while disrupting ordinal color–value correspondence.}
	\label{sequential&randomized}
\end{figure}

This experiment is designed to evaluate whether sequential color ordering influences FM spatial reasoning performance (H2). To this end, we construct two different color-assignment strategies: \textit{Sequential Encoding} and \textit{Randomized Encoding}, as illustrated in Figure \ref{sequential&randomized}. 
\begin{itemize}
    \item \textbf{Sequential Encoding.}  In the sequential setting, we directly adopt the sequential color templates introduced in Sec.~\ref{H1-generated}. Thematic values are mapped to colors following the original ordering defined by the ColorBrewer sequential palettes, thereby preserving monotonic color progression and ordinal consistency.

    \item \textbf{Randomized Encoding.}  In the randomized setting, the same set of colors is retained, but the correspondence between thematic classes and colors is randomly permuted. As a result, the overall color composition remains unchanged, while the ordinal relationship between color progression and thematic magnitude is intentionally disrupted. 
\end{itemize}

\subsubsection{H3: Contrast Manipulation}
\label{H3-generated}

\begin{figure} [H]
	\centering
	\includegraphics[width=\textwidth]{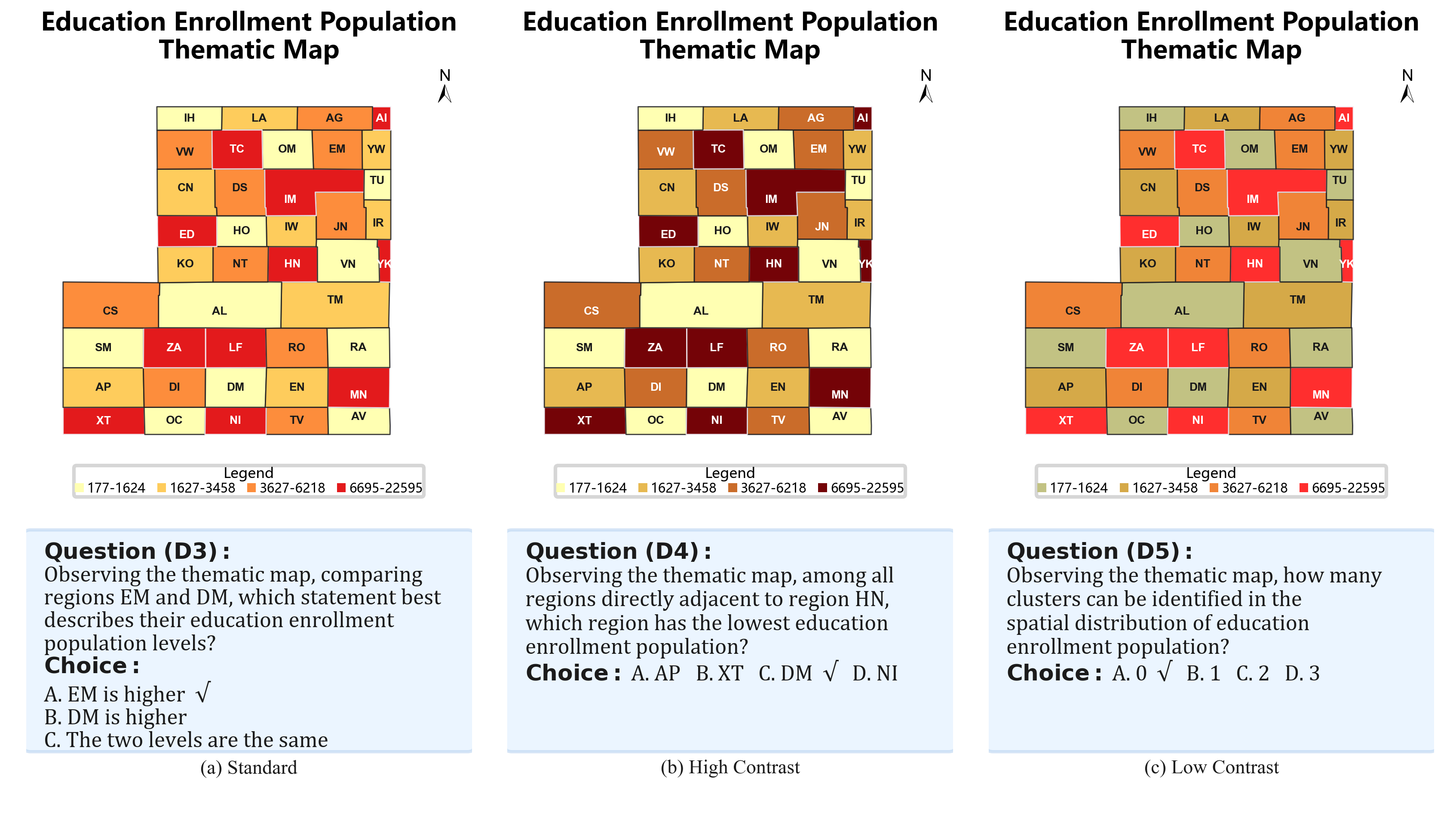}
        \captionsetup{skip=0pt}  % 调整该图的上下间距
	\caption{Examples of standard(a), high-contrast(b), and low-contrast(c) sequential choropleth maps with representative reasoning tasks.}
	\label{chroma}
\end{figure}

This experiment is designed to evaluate whether color contrast influences FM spatial reasoning performance (H3). To this end, we construct three different contrast settings: \textit{Standard Contrast}, \textit{High Contrast}, and \textit{Low Contrast}, as illustrated in Figure \ref{chroma}. We manipulate lightness to increase or decrease contrast, as lightness is the primary visual cue used to represent ordinal relationships in sequential choropleth maps \cite{Brewer03, Slocum2022}. Specifically, adjustments are made in the perceptually uniform CIELab color space rather than the Munsell system used in the original ColorBrewer framework, allowing for precise, continuous, and device-independent control of lightness and color distance \cite{fairchild2002cie}. In CIELab, $L^*$ represents perceptual lightness, while $a^*$ and $b^*$ represent chromatic dimensions. 

\begin{itemize}
    \item \textbf{Standard Contrast.}  In the standard setting, we directly adopt the original sequential color templates introduced in Sec.~\ref{H1-generated}. These palettes preserve the default lightness progression and contrast relationships defined by the ColorBrewer framework.

    \item \textbf{High Contrast.} In the high-contrast setting, the lightness differences between adjacent thematic classes are enlarged by 1.5$\times$ while preserving chromatic dimensions ($a^*$ and $b^*$). Specifically, the brightest end of the sequential palette is used as the anchor point. Let $L_i^*$ denote the lightness value of the $i$-th class and $L_{\max}^*$ denote the maximum lightness value in the palette. The adjusted lightness value $\hat{L}_i^*$ is computed as:

    \begin{equation}
    \hat{L}_i^* = L_{\max}^* - 1.5 \times (L_{\max}^* - L_i^*)
    \end{equation}

    \item \textbf{Low Contrast.}  In the low-contrast setting, the lightness differences between adjacent thematic classes are reduced to 50\% of the original values while preserving chromatic dimensions ($a^*$ and $b^*$). Specifically, the midpoint lightness value between the brightest and darkest classes is first computed as:

    \begin{equation}
    L_{\text{mid}}^* = \frac{L_{\max}^* + L_{\min}^*}{2}
    \end{equation}

    All lightness values are then shifted 50\% toward this midpoint:

    \begin{equation}
    \hat{L}_i^* = L_{\text{mid}}^* + 0.5 \times (L_i^* - L_{\text{mid}}^*)
    \end{equation}

\end{itemize}

To ensure sufficiently distinguishable lightness intervals after contrast manipulation, we only employ 4-class and 5-class sequential schemes in the contrast experiments. Increasing the number of classes would substantially reduce lightness differences between adjacent categories, making controlled high- and low-contrast variations less perceptually distinguishable.

\subsubsection{Map generation}
\label{Map-generated}

Based on the color templates introduced in Sec.~\ref{H1-generated}--Sec.~\ref{H3-generated}, we further construct a fully automated pipeline for thematic map generation. The entire process consists of three major stages: \textbf{geographic region generation}, \textbf{thematic value assignment}, and \textbf{standardized map rendering}.

\textbf{(1) Geographic region generation.}  
The base maps are generated using county-level or equivalent administrative regions from the United States, France, Germany, and Switzerland. Compared with higher-level administrative units, county-level regions contain more complex neighborhood relationships and weaker semantic familiarity, making it more difficult for models to rely on memorized geographic shapes or prior geographic knowledge. To further reduce the influence of memorized geographic outlines, we avoid directly using complete administrative maps with highly recognizable boundaries. Instead, for each benchmark instance, a geographic dataset is randomly selected and reprojected into a unified projected coordinate system. A square clipping window is then randomly placed within the geographic extent, and only subsets containing between 20 and 50 regions are retained.

To ensure topological validity and rendering consistency, all MultiPolygons are exploded into independent Polygon geometries. Invalid geometries are repaired through topology validation operations, and empty geometries are removed. To reduce excessive geometric complexity while preserving adjacency relationships, topology-aware shared-boundary simplification is further applied. Additional geometric quality filters are also used to remove extremely small regions and narrow, fragmented polygons.

\textbf{(2) Thematic value assignment.}  
After the geographic regions are generated, thematic values are assigned according to predefined spatial structures. The underlying attribute values are derived from real-world thematic datasets collected from the MapQA benchmark \cite{chang2022mapqa}. Different spatial structures are incorporated because real-world thematic maps often exhibit distinct spatial organizations, which may substantially influence how FMs perceive and reason about spatial patterns. Following the spatial structure design proposed by Wei et al. (2026) \cite{wei2026mapmatter}, the benchmark includes four representative structures: \textit{cluster}, \textit{trend}, \textit{structure}, and \textit{random}, which are evenly distributed across the benchmark. The detailed definitions and generation procedures of these structures are described in their work. The generated thematic values are subsequently mapped to the color templates introduced in Sec.~\ref{H1-generated}--Sec.~\ref{H3-generated}.

\textbf{(3) Standardized map rendering.}  
Finally, all maps are rendered under standardized cartographic settings. All maps are exported using identical image resolution, polygon boundary width, layout configuration, and background settings. Each map is rendered at 1200 $\times$ 1200 pixels with 220 DPI. Region labels are automatically placed at polygon centroids. To ensure sufficient visual contrast, label and boundary colors are automatically selected according to background lightness. Following standard thematic cartographic conventions, each map additionally includes a legend and a north arrow generated under consistent design specifications to ensure comparable map-reading conditions across all benchmark instances. All maps are generated through a fully automated and reproducible Python pipeline using \texttt{geopandas}, \texttt{matplotlib}.

\subsubsection{Dataset overview}
The benchmark consists of three map groups corresponding to the three hypotheses. For H1 (Hue-controlled), we generate 1,152 maps using 18 sequential palettes under four class settings (4--7 classes), yielding 64 maps for each palette configuration. For H2 (Sequential vs. Randomized Encoding), the sequential condition directly reuses the 1,152 maps from H1, while an additional 1,152 randomized counterparts are generated, resulting in 2,304 maps in total. For H3 (Contrast Manipulation), new maps are generated because the experiment employs only 4-class and 5-class schemes, rather than the 4--7 class settings used in H1 and H2, to preserve sufficient lightness separability after manipulation. Specifically, three contrast settings---\textit{Standard Contrast}, \textit{High Contrast}, and \textit{Low Contrast}---are constructed, each containing 1,152 maps, resulting in 3,456 maps for H3. Overall, the benchmark contains 5,760 choropleth maps.

\subsection{Benchmark Construction}
\label{benchmark}
\subsubsection{Spatial Reasoning Tasks Design}

To systematically evaluate how choropleth color encoding influences FM spatial reasoning, we adopt a multi-level spatial task framework spanning five hierarchical cognitive dimensions: \textit{Attribute Identify}, \textit{Spatial Recognition}, \textit{Compare}, \textit{Rank}, and \textit{Delineate}. The detailed task definitions and question templates are inherited from ChoroplethMap-Bench \cite{wei2026mapmatter}, including 12 fine-grained question types spanning the five task dimensions. The complete task descriptions are summarized in Table~\ref{tab:task-details}. For each generated map, one question is constructed for each task dimension, resulting in five benchmark questions per map (5760 maps and 28800 QA pairs).

\begin{table}[H]
\centering
\scriptsize
\caption{Tasks design, inherited from ChoroplethMap-Bench \cite{wei2026mapmatter}.}
\label{tab:task-details}
\renewcommand{\arraystretch}{1.25}
\setlength{\tabcolsep}{5pt}

\begin{tabular}{>{\centering\arraybackslash}m{0.12\textwidth}
                >{\centering\arraybackslash}m{0.18\textwidth}
                >{\arraybackslash}m{0.66\textwidth}}
\toprule
\textbf{Task dimension} & \textbf{Question Subtype} & \textbf{Question Description} \\
\midrule

\multirow{2}{=}{\centering D1. Attribute Identify}
& attr2region
& Q1. Given an attribute level, identify the corresponding region. \\
& region2attr
& Q2. Given a specific region, identify which attribute level it belongs to. \\
\midrule

\multirow{2}{=}{\centering D2. Spatial Recognition}
& Direction recognition
& Q3. Determine the directional spatial relationship between two regions. \\
& Adjacent recognition
& Q4. Determine whether two regions are adjacent or non-adjacent. \\
\midrule

\centering D3. Compare
& Attribute comparison
& Q5. Compare the attribute values of two given regions and determine which is higher, lower, or equal. \\
\midrule

\multirow{2}{=}{\centering D4. Rank}
& Global rank
& Q6. Find the region with the highest (or lowest) attribute value across the entire map. \\
& Local rank
& Q7. Find the region with the highest (or lowest) attribute value among the neighbors of a given region. \\
\midrule

\multirow{5}{=}{\centering D5. Delineate}
& \multirow{2}{=}{\centering Cluster delineate}
& Q8. Determine the number of spatial clusters. \\
&
& Q9. Determine whether a given region belongs to a cluster. \\

& Trend delineate
& Q10. Identify the direction of a monotonic gradient (trend) in the attribute distribution. \\

& \multirow{2}{=}{\centering Structure delineate}
& Q11. Determine the number of ring structures. \\
&
& Q12. Identify regions located on or inside a ring. \\

\bottomrule
\end{tabular}
\end{table}

\subsubsection{Human validation}

To further verify the reliability and interpretability of the constructed benchmark, we additionally conduct a human validation experiment on a randomly sampled 20\% subset of the benchmark questions (5,760 of 28,800). This pool was then partitioned equally among six participants with no overlap, so that each participant completed 960 questions. Sampling was stratified to ensure that every participant's share spanned sequential and randomized encoding as well as the three lightness-contrast settings. The results are summarized in Table~\ref{tab:human}.

Six participants were recruited from students specializing in Geographic Information Science, providing a level of map-reading familiarity broadly representative of end users of thematic and choropleth maps while remaining independent from the authors' benchmark construction process. For each participant, the presentation order of maps and questions within their assigned subset was independently randomized to control for order and practice effects. Testing was divided into multiple sessions; within each session, participants completed blocks of 50 questions followed by a mandatory rest period of 10--20 minutes to minimize fatigue-related error. The task involved no personally identifiable or sensitive information and posed minimal risk to participants. All participants were informed of the study's purpose and voluntarily agreed to participate.

Overall, human participants achieve consistently high performance across all benchmark settings, with an average overall accuracy of 94.8\%, indicating that the generated maps and task formulations remain highly interpretable for human readers. Under sequential encoding, the average human accuracy reaches 95.7\%, while randomized encoding decreases performance slightly to 91.7\%, suggesting that disrupting ordinal color organization still introduces additional cognitive difficulty during thematic map interpretation. A similar pattern can also be observed in the contrast experiments. Human participants achieve 95.5\% accuracy under the standard condition, 96.7\% under the high-contrast condition, and 94.7\% under the low-contrast condition. The results suggest that increasing lightness contrast slightly improves map readability, whereas reducing contrast weakens visual separability between neighboring thematic classes and introduces moderate performance degradation. Nevertheless, overall human performance remains robust across all benchmark settings.

\begin{table}[H]
\centering
\scriptsize
\caption{Human validation results on a randomly sampled subset of the bench.}
\label{tab:human}
\resizebox{\textwidth}{!}{
\begin{tabular}{lcccccc}
\toprule
Participant & Overall & Sequential & Randomized & Standard & High Contrast & Low Contrast \\
\midrule
P1 & 96.8 & 99.0 & 92.0 & 98.0 & 98.0 & 97.0 \\
P2 & 96.6 & 96.0 & 94.0 & 97.0 & 99.0 & 97.0 \\
P3 & 97.8 & 97.0 & 98.0 & 96.0 & 99.0 & 99.0 \\
P4 & 97.2 & 97.0 & 94.0 & 98.0 & 99.0 & 98.0 \\
P5 & 96.4 & 96.0 & 96.0 & 98.0 & 97.0 & 95.0 \\
P6 & 84.2 & 89.0 & 76.0 & 86.0 & 88.0 & 82.0 \\
\midrule
Avg. & 94.8 & 95.7 & 91.7 & 95.5 & 96.7 & 94.7 \\
\bottomrule
\end{tabular}
}
\end{table}

\subsection{Experimental setting}
\label{evaluation}

\hspace*{1em}\textbf{(1) Evaluated models.} To systematically evaluate the influence of cartographic color encoding on FM spatial reasoning, we evaluate 21 recent multimodal large language models from both open-source and proprietary ecosystems.
The selected models span different model scales, architectures, and vision-language pretraining paradigms. For open-source models, we include representative families such as Qwen3.5, Qwen3-VL, Qwen2.5-VL, InternVL3.5, GLM-4.6V, and Gemma-3. For proprietary models, we evaluate systems including Gemini-3.5-Flash, GPT-5.5, Qwen3.6-Plus, Doubao-Seed-2.0-lite, Kimi-K2.6, ERNIE 5.0, and MiMo-V2.5.

\textbf{(2) Evaluation process.} For each benchmark instance, models receive a choropleth map image together with the corresponding spatial reasoning question. All experiments are conducted under a zero-shot setting. To ensure fair comparison, the same prompting template is used across all models and all map conditions. Each model generates a single answer for the target multiple-choice question. Model outputs are automatically parsed and matched against benchmark ground-truth labels. All models are evaluated independently under identical benchmark settings.

All experiments involving open-source models are conducted on 2-GPU servers equipped with NVIDIA RTX 4090 (24GB) GPUs. To improve large-scale evaluation efficiency, memory-efficient inference settings, including low-bit quantization, are adopted for large vision-language models. The maximum generation length is limited to 64 new tokens. Initial decoding uses deterministic greedy inference to ensure stable outputs across repeated evaluations. If the returned response cannot be parsed into a valid benchmark answer format, an automatic retry mechanism is triggered. Retry decoding uses non-zero sampling parameters to improve response validity, with at most two retries for each failed instance. Proprietary models are evaluated through stable public APIs supporting joint image-text multimodal input under the same automated benchmark pipeline.

\textbf{(3) Evaluation metric.} Model performance is evaluated using accuracy. Accuracy is defined as the proportion of correctly answered instances among all evaluated benchmark questions.

\section{Experimental Results}
Based on the definitions in Sec.~\ref{method}, we investigate three aspects of choropleth color design corresponding to our hypotheses in Sec.~\ref{hypothesis} (H1-H3). All experiments are conducted under identical benchmark settings to ensure that performance differences primarily reflect the influence of color encoding strategies.

\subsection{H1 Hypothesis: Influence of Sequential Hue Palettes}\label{sec:H1-generated}
Table~\ref{tab:hue_effect} summarizes the average accuracy of all evaluated models under 18 sequential palettes in Table \ref{tab:hue_palettes}. From the table, we can have the following observations:

\textbf{(1) Different sequential hue palettes produce only limited and non-systematic performance variation across FMs.} Across all evaluated palettes, the average model accuracy ranges from 51.2\% (\textit{Purples}) to 55.5\% (\textit{YlGn}), corresponding to a maximum global variation of only 4.3 percentage points. Moreover, performance fluctuations remain relatively small and model-dependent, without exhibiting a unified increasing or decreasing trend. To further examine whether the observed hue-related differences are statistically significant, we conducted repeated-measures statistical analysis across different hue conditions. Since the performance distributions satisfied the normality assumption, repeated-measures ANOVA was adopted and indicates an overall significant effect across hue conditions ($p<0.001$). However, post-hoc multiple comparisons using the LSD method were further conducted to examine pairwise differences between individual hue palettes. These results suggest that hue variations contribute only minor, non-systematic fluctuations in model performance, consistent with \textbf{H1}. This conclusion was further confirmed using a generalized linear mixed-effects model (GLMM) analysis reported in Sec.~\ref{sec:glmm-robustness}.

\textbf{(2) Hue-related performance differences vary across model architectures and training paradigms.} Although overall variation remains relatively small, several model-specific tendencies can still be observed. For instance, Qwen3.5-9B performs best under \textit{BuPu} (71.2\%) and \textit{BuGn}/\textit{PuBu} (70.9\%), while Gemini-3.5-Flash achieves its highest accuracy under \textit{YlGnBu} (83.4\%). Similarly, Kimi-K2.6 achieves its highest accuracy under \textit{Blues} (84.4\%) and \textit{YlGnBu} (83.4\%). These variations suggest that hue sensitivity is not fully consistent across architectures and pretraining paradigms. Importantly, however, no single hue palette consistently dominates across all evaluated models. These results indicate that hue-related differences are likely secondary effects rather than dominant determinants of reasoning performance.

\begin{table}[H]
\centering
\caption{Average accuracy (\%) under different sequential hue palettes. Repeated-measures ANOVA: $p<0.001$. Post-hoc multiple comparisons reveal no significant differences between individual hue pairs.}
\label{tab:hue_effect}
\hspace*{-0.1\textwidth}
\resizebox{1.2\textwidth}{!}{
\begin{tabular}{lcccccccccccccccccc}
\toprule
Model & Blues & BuGn & BuPu & GnBu & Greens & Greys & OrRd & Oranges & PuBu & PuBuGn & PuRd & Purples & RdPu & Reds & YlGn & YlGnBu & YlOrBr & YlOrRd\\
\midrule
Qwen3.5-9B & 66.2 & 70.9 & \textbf{71.2} & 67.2 & 69.4 & 68.8 & 68.8 & 68.8 & 70.9 & 70.6 & 68.4 & 66.9 & 65.6 & 69.1 & 67.5 & 68.1 & 67.2 & 68.8\\
Qwen3.5-4B & 61.9 & 60.0 & 64.1 & 63.7 & 62.2 & 62.8 & 62.5 & 60.0 & 59.7 & \textbf{66.6} & 59.4 & 63.7 & 64.7 & 64.7 & 65.0 & 61.3 & 61.3 & 62.8\\
Qwen3.5-2B & 46.9 & 51.2 & \textbf{53.8} & 48.1 & 50.0 & 49.4 & 46.2 & 46.2 & 50.9 & 50.3 & 50.3 & 47.2 & 50.6 & 50.6 & 51.2 & 49.7 & 50.3 & 48.8\\
Qwen3-VL-8B-Instruct & 60.0 & 62.8 & 64.1 & 59.1 & 67.2 & 58.8 & \textbf{69.1} & 60.3 & 61.6 & 62.5 & 57.5 & 58.8 & 62.5 & 65.3 & 60.9 & 64.7 & 61.6 & 61.6\\
Qwen3-VL-4B-Instruct & 55.9 & 59.7 & 62.2 & 56.6 & 62.5 & 55.6 & \textbf{62.5} & 58.4 & 55.9 & 58.4 & 55.0 & 56.9 & 61.3 & 61.6 & 59.7 & 60.3 & 59.1 & 56.9\\
Qwen3-VL-2B-Instruct & 46.9 & 43.8 & 41.6 & 42.2 & 43.8 & 43.8 & 44.1 & 44.1 & 41.2 & 44.7 & 39.1 & 42.5 & 44.1 & \textbf{47.5} & 42.2 & 41.6 & 45.3 & 43.4\\
Qwen2.5-VL-7B-Instruct & 48.4 & 46.6 & 53.4 & 51.2 & 51.6 & 49.4 & 51.2 & 48.8 & 53.8 & \textbf{54.7} & 51.9 & 51.2 & 53.4 & 50.9 & 54.1 & 50.9 & 48.4 & 48.4\\
Qwen2.5-VL-3B-Instruct & 44.7 & 44.7 & 45.9 & 46.2 & 45.9 & 44.4 & 40.9 & 43.1 & 45.0 & 45.6 & 37.8 & 42.5 & 43.4 & 45.3 & \textbf{47.5} & 44.1 & 46.6 & 41.2\\
GLM-4.6V-Flash & \textbf{66.9} & 65.0 & 66.2 & 62.8 & 66.6 & 63.4 & 65.6 & 64.7 & 61.3 & 65.9 & 61.3 & 62.5 & 62.5 & 65.9 & 65.9 & 64.1 & 62.8 & 64.4 \\
Gemma3-12B-IT & 35.6 & 36.2 & 34.1 & 36.9 & 37.2 & 35.9 & 36.9 & 30.9 & 33.1 & 36.9 & 35.6 & 33.8 & 34.4 & 35.9 & \textbf{42.5} & 33.1 & 34.4 & 39.7 \\
Gemma3-4B-IT & 29.1 & 30.0 & 29.1 & \textbf{32.2} & 29.1 & 26.6 & \textbf{32.2} & 28.7 & 30.3 & 29.4 & 29.7 & 31.6 & 30.6 & 27.2 & 30.0 & 26.2 & 30.9 & 30.3 \\
InternVL3.5-8B & 61.9 & 55.9 & 62.5 & 56.2 & 60.9 & 56.9 & 59.1 & 55.3 & 57.2 & 59.7 & 58.1 & 54.4 & 59.7 & 61.6 & 60.0 & \textbf{63.4} & 61.3 & 59.4 \\
InternVL3.5-4B & 58.4 & 55.3 & 62.8 & 58.1 & 59.7 & 54.7 & \textbf{63.7} & 56.9 & 57.8 & 59.7 & 55.0 & 54.4 & 55.6 & 60.9 & 59.4 & 63.1 & 62.5 & 56.2 \\
InternVL3.5-2B & 41.9 & 38.8 & 37.8 & 41.2 & 36.9 & 39.7 & 43.1 & 42.5 & 40.0 & 42.8 & 40.9 & 35.0 & 42.5 & 46.6 & \textbf{47.2} & 40.3 & 43.1 & 40.9 \\
\midrule
Kimi-K2.6 & \textbf{84.4} & 76.2 & 81.2 & 79.7 & 79.7 & 75.6 & 78.4 & 75.6 & 80.0 & 77.2 & 81.2 & 75.0 & 78.7 & 77.5 & 80.9 & 83.4 & 76.9 & 78.1\\
Qwen3.6-plus & 40.3 & 44.1 & 41.9 & 45.3 & 45.3 & \textbf{49.1} & 46.2 & 47.2 & 39.7 & 46.6 & 44.1 & 41.6 & 45.9 & 44.4 & 44.7 & 44.7 & 46.2 & 40.0 \\
Doubao-Seed-2.0-lite & 45.6 & 46.7 & 50.3 & 48.1 & 49.2 & 47.2 & 48.1 & 51.2 & 45.6 & 48.1 & \textbf{52.8} & 43.4 & 46.9 & 46.7 & 48.8 & 51.6 & 43.3 & 46.2 \\
MiMo-V2.5 & 54.1 & 50.0 & 51.6 & 52.2 & \textbf{54.4} & 47.5 & 47.5 & 48.4 & 46.2 & 49.4 & 48.8 & 45.9 & 49.4 & 52.2 & 49.7 & 46.2 & 50.0 & 45.6\\
ERNIE 5.0 & 39.4 & 42.2 & 42.2 & 44.4 & 35.6 & 36.6 & 40.9 & 35.9 & 36.6 & 43.8 & 41.2 & 39.4 & 43.1 & 42.2 & \textbf{45.9} & 43.8 & 40.3 & 33.1\\
Gemini-3.5-Flash & 82.5 & 79.7 & 78.4 & 78.8 & 78.4 & 77.2 & 75.6 & 73.1 & 75.9 & 79.1 & 74.1 & 75.3 & 76.9 & 81.6 & 79.4 & \textbf{83.4} & 76.2 & 76.9 \\
GPT-5.5 & 55.9 & 59.7 & 60.0 & 62.8 & 59.7 & 56.6 & 60.0 & 58.7 & 60.6 & 58.4 & 61.4 & 54.1 & 60.0 & \textbf{63.7} & 62.5 & 62.2 & 55.6 & 62.5\\
\midrule
Avg. & 53.7 & 53.3 & 55.0 & 54.0 & 54.5 & 52.4 & 54.4 & 52.3 & 52.5 & 54.8 & 52.6 & 51.2 & 53.9 & 55.3 & 55.5 & 54.6 & 53.5 & 52.6 \\
\bottomrule
\end{tabular}
}
\end{table}

\subsection{H2 Hypothesis: Sequential versus Randomized Color Encoding}

Table~\ref{tab:sequential_effect} summarizes model performance under sequential and randomized encoding conditions. Contrary to our initial hypothesis (H2), randomized encoding substantially degrades FM reasoning performance, consistent with the human validation results. Across all evaluated models, average accuracy drops from 53.7\% under sequential encoding to 44.5\% under randomized encoding, corresponding to an absolute decline of 9.2 percentage points and a relative performance decrease of 15.9\%. Most models show clear performance losses when sequential ordering is disrupted. For instance, Gemini-3.5-Flash experiences the largest drop, falling from 77.9\% to 58.9\% (-19.0 points), while Qwen3.5-9B and Qwen3-VL-8B-Instruct decline by 15.0 and 14.2 points, respectively. Only a few smaller models, such as Gemma3-4B, remain largely unaffected. A similar trend is observed in human validation: performance decreases from 95.7\% under sequential encoding to 91.7\% under randomized encoding, indicating that stable ordinal color organization also facilitates human interpretation of thematic maps.

To further examine statistical significance, we conducted Shapiro--Wilk normality tests, which confirmed that the performance distributions satisfy the normality assumption. We therefore applied a paired \textit{t}-test to compare sequential and randomized conditions. The results reveal a highly significant difference ($p<0.001$) with a large effect size (Cohen's $d=1.762$), confirming that randomized encoding causes substantial and systematic degradation in FM spatial reasoning performance. These findings indicate that sequential cartographic ordering plays a far more important role in machine map reasoning than suggested by \textbf{H2}. This conclusion is robust to the clustered structure of the benchmark, as revealed by GLMM analysis in Sec.~\ref{sec:glmm-robustness}.

\begin{table}[H]
\centering
\scriptsize
\caption{Performance comparison between sequential and randomized color encoding. $\Delta$ indicates the accuracy drop after disrupting sequential ordering. Paired T-Test: $p<0.001$, Cohen's $d=1.762$.}
\label{tab:sequential_effect}
\resizebox{\textwidth}{!}{
\begin{tabular}{lcccc}
\toprule
Model & Sequential (\%) & Randomized (\%) & $\Delta$ (\%) & Relative Drop (\%) \\
\midrule
Qwen3.5-9B & \textbf{68.6} & 53.6 & -15.0 & -21.9 \\
Qwen3.5-4B & \textbf{62.6} & 49.1 & -13.5 & -21.6 \\
Qwen3.5-2B & \textbf{49.5} & 41.4 & -8.1 & -16.4 \\
Qwen3-VL-8B-Instruct & \textbf{62.1} & 48.0 & -14.2 & -22.7 \\
Qwen3-VL-4B-Instruct & \textbf{58.8} & 45.8 & -13.0 & -22.1 \\
Qwen3-VL-2B-Instruct & \textbf{43.4} & 36.3 & -7.1 & -16.4 \\
Qwen2.5-VL-7B-Instruct & \textbf{51.0} & 41.6 & -9.4 & -18.4 \\
Qwen2.5-VL-3B-Instruct & \textbf{44.2} & 38.0 & -6.2 & -14.0 \\
GLM-4.6V-Flash & \textbf{64.3} & 51.9 & -12.4 & -19.3 \\
Gemma3-12B-IT & \textbf{35.7} & 35.3 & -0.4 & -1.1 \\
Gemma3-4B-IT & 29.6 & \textbf{29.7} & +0.1 & +0.3 \\
InternVL3.5-8B & \textbf{59.1} & 46.3 & -12.8 & -21.7 \\
InternVL3.5-4B & \textbf{58.6} & 47.0 & -11.6 & -19.8 \\
InternVL3.5-2B & \textbf{41.2} & 35.4 & -5.8 & -14.1 \\
\midrule
Kimi-K2.6 & \textbf{78.9} & 65.1 & -13.8 & -17.5 \\
Qwen3.6-plus & \textbf{44.3} & 41.1 & -3.2 & -7.2 \\
Doubao-Seed-2.0-lite & \textbf{47.8} & 44.9 & -2.9 & -6.1 \\
MiMo-V2.5 & \textbf{49.4} & 38.3 & -11.1 & -22.5 \\
ERNIE 5.0 & \textbf{40.4} & 37.2 & -3.2 & -7.9 \\
Gemini-3.5-Flash & \textbf{77.9} & 58.9 & -19 & -24.4 \\
GPT-5.5 & \textbf{59.7} & 49.2 & -10.5 & -17.6 \\
\midrule
Avg. & \textbf{53.7} & 44.5 & -9.2 & -15.9 \\
\midrule
Human & \textbf{95.7} & 91.7 & -4.0 & -4.2 \\
\bottomrule
\end{tabular}
}
\end{table}

\begin{figure}[H]
\centering
\includegraphics[width=\textwidth]{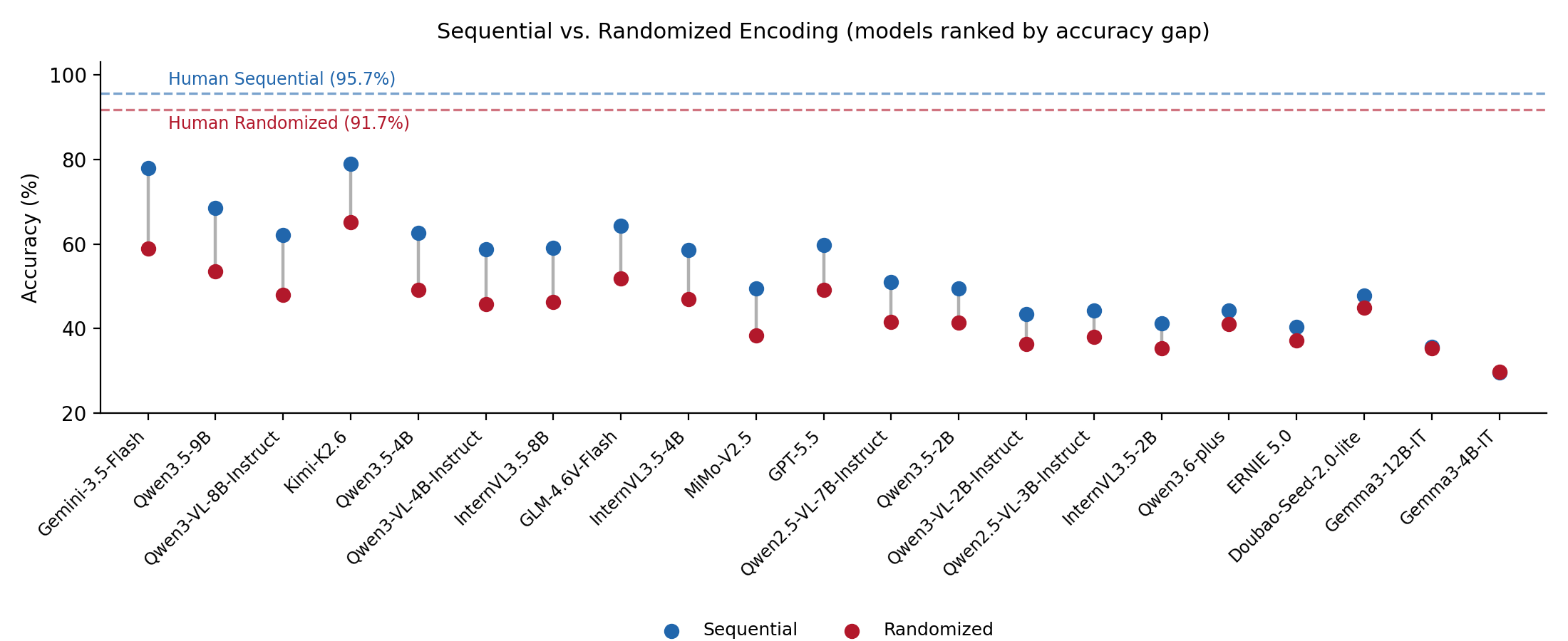}
\caption{Per-model accuracy under sequential vs.\ randomized color encoding. Vertical line segments indicate the magnitude of degradation; models are ordered by gap size (largest at left). Dashed horizontal lines show human accuracy under each condition for reference.}
\label{fig:h2_dumbbell}
\end{figure}

Figure~\ref{fig:h2_dumbbell} further visualizes this comparison directly: while every evaluated model falls well below the human accuracy lines under both conditions, the vertical gap between each model's sequential and randomized accuracy is, for most models, considerably larger than the corresponding human gap (4.0 percentage points), indicating that FMs are, in relative terms, more disrupted by randomized color ordering than human readers are.

\subsection{H3 Hypothesis: Influence of Lightness Contrast}\label{sec:H3-generated}

Table~\ref{tab:luminance_effect} summarizes model performance under different lightness contrast conditions. Overall, lightness contrast systematically influences FM spatial reasoning performance across most evaluated models. Compared with the standard setting, increasing contrast slightly improves the average accuracy from 55.3\% to 55.8\% (+0.5), whereas reducing contrast decreases it to 53.3\% (-2.0). Correspondingly, 17 out of 21 models improve under the high-contrast condition, while 19 out of 21 models degrade under the low-contrast condition. Similar patterns can be observed across different model families. For example, Qwen3.5-9B and GLM-4.6V-Flash exhibit only marginal gains under enhanced contrast (+0.1 for both), yet show clear degradation under reduced contrast (-3.2 and -3.8, respectively). Comparable trends are also observed in Gemini-3.5-Flash, GPT-5.5, and InternVL3.5. To verify statistical significance, we conducted Friedman tests across contrast conditions. The results reveal a significant overall effect ($p<0.001$) with a small effect size (Cohen's $f=0.084$). Post-hoc comparisons further show significant differences between the Standard and Low Contrast conditions ($p=0.004$, Cohen's $d=0.152$), as well as between the High and Low Contrast conditions ($p<0.001$, Cohen's $d=0.193$). In contrast, the model-level Friedman post-hoc comparison did not detect a statistically significant difference between the Standard and High Contrast conditions, consistent with a diminishing benefit of additional contrast enhancement once sufficient separability is achieved. These results partially support \textbf{H3}. Increasing lightness contrast does improve model performance, but the improvement becomes relatively limited once contrast exceeds a certain separability threshold. In contrast, compressing lightness differences consistently produces substantial performance degradation. This asymmetric pattern suggests that FMs strongly depend on sufficient visual separability between neighboring thematic classes, while excessive additional contrast contributes relatively little new information once ordinal relationships are already clearly distinguishable.

Human validation exhibits a related but notably different pattern. Human accuracy increases from 95.5\% under the standard condition to 96.7\% under the high-contrast condition, and decreases to 94.7\% under the low-contrast condition. Unlike FMs, humans continue to benefit from additional contrast even when baseline performance is already very high. This suggests a fundamental difference between human and machine map interpretation: FMs mainly require sufficient low-level visual separability to support reasoning, whereas human readers can continue to exploit enhanced perceptual contrast even beyond this threshold.

\begin{table}[H]
\centering
\scriptsize

\caption{Performance comparison under different luminance contrast conditions. $\Delta_{L+}$ and $\Delta_{L-}$ indicate the performance change relative to the standard luminance condition. Friedman test results: $p<0.001$, Cohen's $f=0.084$. Post-hoc multiple comparisons: Standard vs. Low Contrast ($p=0.004$, Cohen's $d=0.152$), High vs. Low Contrast ($p<0.001$, Cohen's $d=0.193$).}
\label{tab:luminance_effect}
\resizebox{\textwidth}{!}{
\begin{tabular}{lccccc}
\toprule
Model & Standard (\%) & High Contrast (\%) & $\Delta_{L+}$ (\%) & Low Contrast (\%) & $\Delta_{L-}$ (\%) \\
\midrule
Qwen3.5-9B & 70.8 & \textbf{70.9} & +0.1 & 67.6 & -3.2 \\
Qwen3.5-4B & \textbf{63.6} & \textbf{63.6} & 0.0 & 60.2 & -3.4 \\
Qwen3.5-2B & 51.2 & \textbf{52.0} & +0.8 & 49.1 & -2.1 \\
Qwen3-VL-8B-Instruct & 63.3 & \textbf{64.1} & +0.8 & 60.3 & -3.0 \\
Qwen3-VL-4B-Instruct & 59.7 & \textbf{60.2} & +0.5 & 56.7 & -3.0 \\
Qwen3-VL-2B-Instruct & 44.5 & \textbf{44.7} & +0.2 & 42.6 & -1.9 \\
Qwen2.5-VL-7B-Instruct & 52.2 & \textbf{53.1} & +0.9 & 50.1 & -2.1 \\
Qwen2.5-VL-3B-Instruct & 46.2 & \textbf{47.4} & +1.2 & 45.1 & -1.1 \\
GLM-4.6V-Flash & 66.5 & \textbf{66.6} & +0.1 & 62.7 & -3.8 \\
Gemma-3-12B-IT & \textbf{36.8} & 36.7 & -0.1 & 36.2 & -0.6 \\
Gemma-3-4B-IT & 29.6 & 29.4 & -0.2 & \textbf{29.8} & +0.2 \\
InternVL3.5-8B & 62.2 & \textbf{64.2} & +2.0 & 60.4 & -1.8 \\
InternVL3.5-4B & 59.9 & \textbf{60.9} & +1.0 & 57.0 & -2.9 \\
InternVL3.5-2B & 41.0 & \textbf{41.9} & +0.9 & 40.2 & -0.8 \\
\midrule
Kimi-K2.6 & 80.9 & \textbf{81.8} & +0.9 & 79.4 & -1.5 \\
Qwen3.6-plus & 46.9 & \textbf{47.7} & +0.8 & 44.7 & -2.2 \\
Doubao-Seed-2.0-lite & 49.5 & \textbf{50.1} & +0.6 & 49.0 & -0.5 \\
MiMo-V2.5 & 50.7 & \textbf{51.3} & +0.6 & 49.4 & -1.3 \\
ERNIE 5.0 & \textbf{43.3} & 43.1 & -0.2 & 43.2 & -0.1 \\
Gemini-3.5-Flash & 80.4 & \textbf{80.5} & +0.1 & 76.7 & -3.7 \\
GPT-5.5 & 61.4 & \textbf{61.6} & +0.2 & 58.6 & -2.8 \\
\midrule
Avg. & 55.3 & \textbf{55.8} & +0.5 & 53.3 & -2.0 \\
\midrule
Human & 95.5 & \textbf{96.7} & +1.2 & 94.7 & -0.8 \\
\bottomrule
\end{tabular}
}
\end{table}

\begin{figure}[H]
\centering
\includegraphics[width=\textwidth]{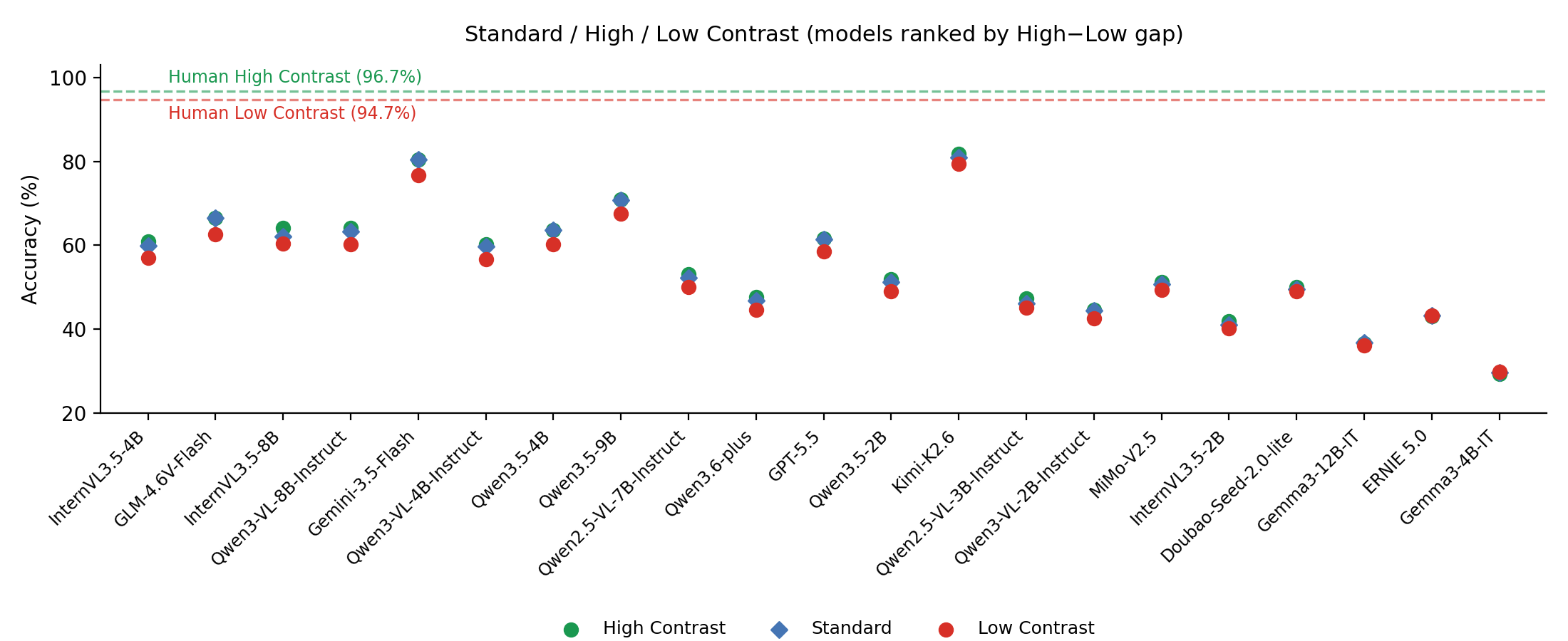}
\caption{Per-model accuracy under Standard, High Contrast, and Low Contrast conditions, ordered by the High-to-Low range (largest at left). Standard and High Contrast markers nearly overlap for most models, illustrating diminishing returns from contrast enhancement, while Low Contrast is consistently pulled downward. Dashed horizontal lines show human accuracy for reference.}
\label{fig:h3_dumbbell}
\end{figure}

Figure~\ref{fig:h3_dumbbell} illustrates this contrast between humans and models: for nearly every model, the Standard and High Contrast markers nearly coincide, while the Low Contrast marker is visibly pulled downward — the same asymmetric pattern seen in the human reference lines, though the relative size of the Low-Contrast drop is generally larger for models than for humans (whose High-to-Low gap remains comparatively narrow, 2.0 percentage points).

\subsection{Robustness to Clustered Data via Generalized Linear Mixed-Effects Models}
\label{sec:glmm-robustness}

The analyses reported above treat model-level and dimension-level accuracies as independent observations, whereas questions are nested within maps and models are nested within families (e.g., Qwen3.5-9B/4B/2B). To verify that none of our reported effects reflect underestimated standard errors due to this clustering, we re-fit each comparison as a generalized linear mixed-effects model (GLMM) at the individual question level (binomial link), with crossed random intercepts for base map and random intercepts for model nested within model family. Models were fit using \texttt{glmmTMB}; Tukey-adjusted pairwise contrasts were obtained via \texttt{emmeans}. Table~\ref{tab:glmm-summary} summarizes all re-analyses, including task-dimension breakdowns discussed further in Secs.~\ref{sec:H2-task-dependent} and~\ref{sec:H3-task-dependent}.

\textbf{H1 (Hue).} No pairwise contrast between any of the 17 hue palettes and the reference palette (\textit{Blues}) reached significance (all $|z|<1.32$, all $p>.18$), confirming that the limited hue effect reported in Sec.~\ref{sec:H1-generated} is not an artifact of clustering.

\textbf{H2 (Sequential vs.\ Randomized Ordering).} Randomized encoding remained a highly significant predictor of reduced accuracy (odds ratio $=1.49$, $z=46.96$, $p<0.001$; predicted accuracy 54.7\% vs.\ 44.7\%), closely matching the aggregate gap in Table~\ref{tab:sequential_effect}. A more conservative model allowing model-family-specific random slopes for condition — to absorb the heterogeneity in models' sensitivity evident in Table~\ref{tab:sequential_effect} — fit the data better ($\mathrm{AIC}=315{,}499$ vs.\ $316{,}162$) and increased the standard error of the fixed effect nearly eight-fold (SE: $0.0085\rightarrow0.0669$), yet the effect remained significant ($z=5.90$, $p<0.001$). A model including a random intercept for country of origin yielded essentially identical estimates.

\textbf{H3 (Lightness Contrast).} Relative to Standard, Low Contrast remained significantly worse (odds ratio $=0.91$, $z=-10.44$, $p<0.001$), while High Contrast showed a small but, at this much larger sample size, statistically detectable improvement (odds ratio $=1.02$, $z=2.77$, $p=0.016$) — an effect an order of magnitude smaller than the Low-Contrast degradation. This refines rather than contradicts Sec.~\ref{sec:H3-generated}: the model-level Friedman test lacked power to detect the Standard-to-High-Contrast difference, but the question-level analysis ($N$\,=\,362, 640) did.

Across H1--H3, the direction and relative magnitude of all reported effects were reproduced after accounting for the non-independence of questions nested within maps and models nested within families. Task-dimension-specific robustness checks are reported alongside the corresponding discussions in Secs.~\ref{sec:H2-task-dependent} and~\ref{sec:H3-task-dependent}.

\begin{table}[H]
\centering
\scriptsize
\caption{Summary of GLMM robustness analyses at the individual question level. All models include crossed random intercepts for the base map and random intercepts for models nested within model family; the H2 random-slope model additionally includes model-family-specific random slopes for condition. For all rows, odds ratio $>1$ indicates higher accuracy under the first-named condition, and $z$ is signed accordingly.}
\label{tab:glmm-summary}
\begin{tabular}{llccc}
\toprule
Hypothesis & Contrast & Odds Ratio & $z$ & $p$ \\
\midrule
H1 (Hue) & 17 palettes vs.\ Blues & — & $|z|<1.32$ & all $>.18$ \\
\midrule
H2 (main) & Sequential vs.\ Randomized & 1.49 & 46.96 & $<.001$ \\
H2 (random slope) & Sequential vs.\ Randomized & 1.48 & 5.90 & $<.001$ \\
H2 (+ country) & Sequential vs.\ Randomized & 1.49 & 46.96 & $<.001$ \\
H2 $\times$ D1 & Sequential vs.\ Randomized & 1.24 & 10.12 & $<.001$ \\
H2 $\times$ D2 & Sequential vs.\ Randomized & 1.01 & 0.45 & .656 \\
H2 $\times$ D3 & Sequential vs.\ Randomized & 2.06 & 32.51 & $<.001$ \\
H2 $\times$ D4 & Sequential vs.\ Randomized & 3.62 & 60.02 & $<.001$ \\
H2 $\times$ D5 & Sequential vs.\ Randomized & 1.25 & 9.99 & $<.001$ \\
\midrule
H3 (main) & High vs.\ Standard & 1.02 & 2.77 & .016 \\
H3 (main) & Low vs.\ Standard & 0.91 & $-10.44$ & $<.001$ \\
H3 (main) & High vs.\ Low & 1.12 & 13.21 & $<.001$ \\
H3 $\times$ D1 & High vs.\ Standard & 1.23 & 8.65 & $<.001$ \\
H3 $\times$ D1 & Low vs.\ Standard & 0.78 & $-11.05$ & $<.001$ \\
H3 $\times$ D2 & High vs.\ Standard & 1.01 & 0.36 & .930 \\
H3 $\times$ D2 & Low vs.\ Standard & 1.04 & 1.74 & .192 \\
H3 $\times$ D3 & High vs.\ Standard & 0.98 & $-0.67$ & .784 \\
H3 $\times$ D3 & Low vs.\ Standard & 0.90 & $-4.32$ & $<.001$ \\
H3 $\times$ D4 & High vs.\ Standard & 0.99 & $-0.31$ & .947 \\
H3 $\times$ D4 & Low vs.\ Standard & 0.78 & $-11.49$ & $<.001$ \\
H3 $\times$ D5 & High vs.\ Standard & 0.99 & $-0.51$ & .865 \\
H3 $\times$ D5 & Low vs.\ Standard & 0.96 & $-1.94$ & .129 \\
\bottomrule
\end{tabular}
\end{table}
\section{Discussion}

\subsection{Sequential Ordering Influences Thematic Reasoning More Than Pure Spatial Recognition}\label{sec:H2-task-dependent}

To further understand how sequential cartographic ordering affects different forms of spatial reasoning, we additionally compare sequential and randomized encoding across the five task dimensions (Table~\ref{tab:seq_task}). The results reveal a clear task-dependent effect. Sequential ordering substantially influences tasks involving thematic magnitude comparison, while exerting relatively limited influence on pure spatial recognition tasks.

Among all task dimensions, D4 exhibits the strongest degradation after disrupting sequential ordering. Across all evaluated models, the average accuracy decreases dramatically from 58.4\% under sequential encoding to 33.8\% under randomized encoding, corresponding to an absolute degradation of 24.6 percentage points. D3 also shows substantial degradation, decreasing from 49.2\% to 36.7\% (-12.5). In comparison, D1 and D5 exhibit more moderate performance declines, whereas D2 remains almost unchanged (59.3\% vs. 59.2\%). Similar trends can also be observed in the human validation results. Human performance under D3 decreases from 96.7\% to 86.7\%, while D4 decreases from 95.0\% to 90.0\%, indicating that sequential ordering also facilitates human thematic comparison and ranking tasks. In contrast, D5 remains unchanged (92.5\% vs. 92.5\%), and D2 only exhibits minor variation (95.0\% vs. 92.5\%). To further verify the statistical significance of these observations, we conducted significance analyses separately for each task dimension. Since D1, D3, and D5 do not satisfy the normality assumption, Wilcoxon Signed-Rank Tests were adopted, whereas paired T-Tests were used for D2 and D4. The results reveal significant differences for D1 ($p<0.001$, Cohen's $d=0.217$), D3 ($p<0.001$, Cohen's $d=1.287$), D4 ($p<0.001$, Cohen's $d=1.843$), and D5 ($p<0.001$, Cohen's $d=0.98$). In contrast, D2 shows no statistically significant difference ($p=0.476$). The effect sizes further indicate that sequential ordering exerts a particularly strong influence on D3 and D4.

A GLMM with a condition $\times$ task-dimension interaction, accounting for the clustering of questions within maps and models within families (Sec.~\ref{sec:glmm-robustness}), reproduced the same ranking of sensitivity: D4 (odds ratio $=3.62$) $>$ D3 (2.06) $>$ D1 (1.24) $\approx$ D5 (1.25) $>$ D2 (1.01, $p=0.656$, null), with the interaction term itself significant ($p<0.001$) and all non-null contrasts significant at $p<0.001$. This confirms that the task-dependent pattern is not an artifact of the benchmark's clustering structure. Full results (Q1--Q12) across all 21 evaluated models are provided in Appendix~\ref{app:subtype} (Fig.~\ref{fig:h2_subtype_heatmap}).

This hierarchy of sensitivity is highly consistent with the cognitive characteristics of the five task dimensions. Both D3 and D4 strongly rely on relative thematic magnitude comparison and ordinal relationship inference. Sequential choropleth encoding provides a stable monotonic correspondence between lightness progression and attribute magnitude, enabling models to more easily compare neighboring regions and infer regional hierarchies. Once this ordinal organization is disrupted by randomized encoding, the visual relationship between regions becomes substantially less interpretable, greatly increasing the difficulty of comparison-based thematic reasoning. In contrast, D1 mainly focuses on local thematic decoding, while D5 involves higher-level spatial structure perception that depends only partially on sequential ordering. D2, however, primarily relies on geometric and topological spatial relationships, such as adjacency and directional recognition, which are largely independent of thematic magnitude interpretation. Consequently, disrupting sequential lightness organization produces almost no observable influence on D2 performance.

More broadly, these results indicate that FMs do not merely rely on isolated local color tokens during map understanding. Instead, similar to human readers, they appear to exploit the conventional sequential visual organization embedded in sequential choropleth maps to support higher-level comparison and thematic reasoning. Interestingly, this similarity suggests that part of human map interpretation may also rely on learned visual and spatial patterns acquired through experience, rather than relying purely on isolated low-level visual features or token-level color recognition.

\begin{table}[H]
\centering
\scriptsize
\caption{Performance comparison between sequential and randomized encoding across different task dimensions. Wilcoxon Signed-Rank Test, D1 ($p<0.001$, Cohen's $d=0.217$), D3 ($p<0.001$, Cohen's $d=1.287$),  D4 ($p<0.001$, Cohen's $d=1.843$). Paired T-Test: D2 ($p=0.476$, Cohen's $d=0.16$), D5 ($p<0.001$, Cohen's $d=0.98$).}
\label{tab:seq_task}
\resizebox{\textwidth}{!}{
\begin{tabular}{lcccccccccc}
\toprule
& \multicolumn{2}{c}{D1}
& \multicolumn{2}{c}{D2}
& \multicolumn{2}{c}{D3}
& \multicolumn{2}{c}{D4}
& \multicolumn{2}{c}{D5} \\
\cmidrule(lr){2-3}
\cmidrule(lr){4-5}
\cmidrule(lr){6-7}
\cmidrule(lr){8-9}
\cmidrule(lr){10-11}
Model
& Seq. & Rand.
& Seq. & Rand.
& Seq. & Rand.
& Seq. & Rand.
& Seq. & Rand. \\
\midrule

Qwen3.5-9B
& \textbf{84.8} & 80.3
& 75.2 & \textbf{75.9}
& \textbf{59.7} & 35.3
& \textbf{76.0} & 37.8
& \textbf{47.2} & 38.7 \\
Qwen3.5-4B
& \textbf{74.8} & 71.4
& \textbf{70.5} & 68.8
& \textbf{55.9} & 33.6
& \textbf{66.9} & 33.4
& \textbf{44.7} & 38.5 \\
Qwen3.5-2B
& \textbf{64.5} & 60.8
& \textbf{56.4} & 55.9
& \textbf{42.4} & 35.2
& \textbf{56.5} & 28.0
& \textbf{28.0} & 27.0 \\
Qwen3-VL-8B-Instruct
& \textbf{75.4} & 68.5
& \textbf{73.2} & 72.5
& \textbf{54.4} & 32.9
& \textbf{75.1} & 34.7
& \textbf{32.5} & 31.6 \\
Qwen3-VL-4B-Instruct
& \textbf{66.7} & 62.7
& 66.2 & \textbf{66.7}
& \textbf{50.9} & 32.0
& \textbf{72.2} & 33.1
& \textbf{38.0} & 34.4 \\
Qwen3-VL-2B-Instruct
& \textbf{43.5} & 40.9
& 45.6 & \textbf{46.4}
& \textbf{40.6} & 33.0
& \textbf{51.6} & 49.2
& \textbf{35.9} & 32.1 \\
Qwen2.5-VL-7B-Instruct
& \textbf{59.8} & 55.2
& \textbf{60.6} & 58.8
& \textbf{44.2} & 36.0
& \textbf{60.4} & 31.2
& \textbf{30.1} & 26.6 \\
Qwen2.5-VL-3B-Instruct
& \textbf{55.4} & 53.8
& 42.4 & \textbf{43.6}
& \textbf{39.3} & 34.8
& \textbf{55.6} & 31.9
& \textbf{28.1} & 26.0 \\
GLM-4.6V-Flash
& \textbf{81.6} & 81.2
& 66.1 & \textbf{66.9}
& \textbf{55.9} & 37.9
& \textbf{76.8} & 40.1
& \textbf{41.1} & 33.1 \\
Gemma3-12B-IT
& 23.8 & \textbf{24.4}
& 58.9 & \textbf{60.0}
& \textbf{35.3} & 34.2
& \textbf{27.6} & 25.4
& \textbf{32.9} & 32.1 \\
Gemma3-4B-IT
& 25.4 & \textbf{25.6}
& \textbf{32.9} & 32.7
& \textbf{33.4} & 32.8
& \textbf{26.7} & 25.9
& 29.6 & \textbf{31.4} \\
InternVL3.5-8B
& \textbf{74.0} & 72.0
& \textbf{65.1} & 64.4
& \textbf{51.0} & 33.2
& \textbf{75.5} & 36.2
& \textbf{29.8} & 25.6 \\
InternVL3.5-4B
& \textbf{73.7} & 73.1
& \textbf{61.4} & 59.5
& \textbf{55.8} & 35.2
& \textbf{70.9} & 34.6
& 31.2 & \textbf{32.6} \\
InternVL3.5-2B
& \textbf{44.4} & 42.4
& \textbf{43.3} & 43.0
& \textbf{39.9} & 34.3
& \textbf{45.9} & 27.3
& \textbf{32.4} & 30.1 \\
\midrule
Kimi-K2.6
& \textbf{88.6} & 88.1
& 87.4 & \textbf{87.6}
& \textbf{77.5} & 57.5 
& \textbf{86.6} & 49.3
& \textbf{54.4} & 43.1 \\
Qwen3.6-plus
& \textbf{65.5} & 59.2
& 45.2 & \textbf{46.1}
& \textbf{37.8} & 35.2
& \textbf{38.3} & 30.2
& 34.6 & \textbf{34.8} \\
Doubao-Seed-2.0-lite
& 78.0 & \textbf{79.2}
& 49.0 & \textbf{49.6}
& \textbf{37.8} & 34.3
& \textbf{44.4} & 34.9
& \textbf{29.6} & 26.4 \\
MiMo-V2.5
& \textbf{61.6} & 50.0
& \textbf{48.4} & 45.8
& \textbf{48.7} & 33.8
& \textbf{52.7} & 28.4
& \textbf{35.6} & 33.3 \\
ERNIE 5.0
& \textbf{48.7} & 42.7
& 48.9 & \textbf{49.5}
& \textbf{38.1} & 35.3
& \textbf{34.8} & 28.5
& \textbf{31.4} & 30.0 \\
Gemini-3.5-Flash
& \textbf{86.6} & 71.0
& \textbf{80.0} & 79.1
& \textbf{74.7} & 50.0
& \textbf{79.9} & 37.8
& \textbf{68.4} & 56.4 \\
GPT-5.5
& \textbf{68.9} & 60.8
& 69.6 & \textbf{69.8}
& \textbf{59.5} & 43.6
& \textbf{52.0} & 31.0
& \textbf{48.4} & 40.6 \\
\midrule
Avg.
& \textbf{64.1} & 60.2
& \textbf{59.3} & 59.2
& \textbf{49.2} & 36.7
& \textbf{58.4} & 33.8
& \textbf{37.3} & 33.5 \\
\midrule
Human
& \textbf{99.2} & 96.7
& \textbf{95.0} & 92.5
& \textbf{96.7} & 86.7
& \textbf{95.0} & 90.0
& \textbf{92.5} & \textbf{92.5} \\

\bottomrule
\end{tabular}
}
\end{table}

\subsection{Lightness Contrast Mainly Influences Attribute Identification}\label{sec:H3-task-dependent}

To further understand how lightness contrast affects different forms of spatial reasoning, we additionally compare standard, high-contrast, and low-contrast conditions across the five task dimensions (Table~\ref{tab:contrast_task}). The results reveal a clear task-dependent pattern. Lightness contrast manipulation primarily influences tasks that rely on local attribute identification and regional separability, while exerting relatively limited influence on pure spatial recognition and higher-level structural reasoning tasks.

Among all task dimensions, D1 exhibits the strongest sensitivity to lightness contrast manipulation. Across all evaluated models, the average accuracy increases from 68.2\% under the standard condition to 71.2\% under the high-contrast condition, while decreasing substantially to 64.2\% under the low-contrast condition. This corresponds to an average improvement of 3.0 percentage points under enhanced contrast and a degradation of 4.0 percentage points under compressed contrast. In comparison, D4 shows moderate degradation under reduced contrast (60.4\% vs. 55.9\%), whereas D3 and D5 exhibit relatively limited changes. D2 remains almost entirely unaffected across all contrast settings (58.7\%, 58.9\%, and 59.4\%). In contrast to FMs, human performance exhibits a noticeably different sensitivity pattern across task dimensions. For D1, human accuracy remains near ceiling level across all contrast settings (100.0\%, 98.3\%, and 99.2\%), suggesting that local attribute identification is already relatively easy for human readers once basic thematic separability is preserved. However, unlike FMs, human performance under D2, D3, and D5 generally improves as lightness contrast increases. For example, D2 increases from 95.8\% under the standard condition to 97.5\% under the high-contrast condition, D3 increases from 96.7\% to 98.3\%, and D5 increases from 90.0\% to 94.2\%. To further verify the statistical significance of these observations, we conducted significance analyses separately for each task dimension. Since D1, D3, and D5 do not satisfy the normality assumption, Friedman tests were adopted, whereas repeated-measures ANOVA was used for D2 and D4. The results reveal significant differences for D1 ($p<0.001$, Cohen's $f=0.153$) and D3 ($p=0.007$, Cohen's $f=0.056$), while D5 shows no statistically significant difference ($p=0.229$). For D2 ($p=0.024$) and D4 ($p<0.001$), repeated-measures ANOVA indicates overall significance, although post-hoc multiple comparisons reveal no statistically significant differences between individual contrast conditions. Among all task dimensions, D1 exhibits the largest effect size, further confirming that lightness contrast primarily influences local thematic identification. These results suggest that humans can continue benefiting from enhanced contrast even when baseline performance is already very high. In comparison, FMs mainly benefit from contrast enhancement in tasks strongly dependent on local luminance separability, while additional contrast provides relatively limited gains once sufficient visual distinguishability has been established. One possible explanation is that human readers are able to further exploit higher-level visual and spatial organizational patterns under enhanced contrast, whereas FMs still rely more heavily on low-level luminance separability during thematic reasoning.

A clustering-robust GLMM with a condition $\times$ task-dimension interaction, accounting for the nesting of questions within maps and models within families (Sec.~\ref{sec:glmm-robustness}), reproduced this pattern: only D1 showed significant, bidirectional sensitivity to contrast manipulation (High vs. Standard: odds ratio $=1.23$; Low vs. Standard: odds ratio $=0.78$; both $p<0.001$); D3 and D4 degraded significantly only under Low Contrast (odds ratio $=0.90$ and $0.78$, respectively; both $p<0.001$) but showed no significant change under High Contrast (both $p>0.5$); D2 and D5 showed no significant differences under any contrast condition (all $p>0.13$). This confirms that the dimension-specific ordering of sensitivity to lightness contrast reported above is not an artifact of the benchmark's clustering structure. Full results (Q1--Q12) across all 21 evaluated models are provided in Appendix~\ref{app:subtype} (Fig.~\ref{fig:h3_subtype_heatmap}).

More broadly, these findings reveal an important difference between human and machine map understanding. For FMs, lightness contrast mainly facilitates local thematic identification and regional separability, with performance gains becoming limited once sufficient visual distinguishability has been established. Human readers, however, can continue benefiting from enhanced contrast even under already high baseline accuracy, suggesting that they further exploit higher-level visual and spatial organizational patterns beyond low-level luminance separability. Compared with hue variation, lightness contrast therefore constitutes a substantially more fundamental visual signal for FM choropleth understanding, while playing a broader perceptual facilitation role in human spatial cognition.

\begin{table}[H]
\centering
\scriptsize
\caption{Performance comparison under different luminance contrast conditions across task dimensions. Friedman test results: D1 ($p<0.001$, Cohen's $f=0.153$), D3 ($p=0.007$, Cohen's $f=0.056$), and D5 ($p=0.229$, Cohen's $f=0.03$). Repeated-measures ANOVA results: D2 ($p=0.024$), D4 ($p<0.001$); post-hoc multiple comparisons reveal no significant differences in D2 and D4.}
\label{tab:contrast_task}
\resizebox{\textwidth}{!}{
\begin{tabular}{lccccccccccccccc}
\toprule
& \multicolumn{3}{c}{D1}
& \multicolumn{3}{c}{D2}
& \multicolumn{3}{c}{D3}
& \multicolumn{3}{c}{D4}
& \multicolumn{3}{c}{D5} \\
\cmidrule(lr){2-4}
\cmidrule(lr){5-7}
\cmidrule(lr){8-10}
\cmidrule(lr){11-13}
\cmidrule(lr){14-16}
Model
& Std. & L+ & L-
& Std. & L+ & L-
& Std. & L+ & L-
& Std. & L+ & L-
& Std. & L+ & L- \\
\midrule
Qwen3.5-9B
& 90.3 & \textbf{93.0} & 85.0
& \textbf{74.3} & 74.1 & 73.4
& \textbf{61.4} & 60.9 & 59.5
& \textbf{79.8} & 78.1 & 73.9
& 48.2 & \textbf{48.4} & 46.2 \\
Qwen3.5-4B
& 79.5 & \textbf{82.3} & 75.0
& 65.6 & \textbf{66.1} & 65.9
& \textbf{58.7} & 57.0 & 54.9
& \textbf{68.4} & 67.4 & 61.5
& \textbf{45.8} & 45.2 & 43.6 \\
Qwen3.5-2B
& 64.3 & \textbf{69.0} & 60.7
& 58.2 & 57.3 & \textbf{58.6}
& 43.2 & 43.3 & \textbf{44.2}
& 60.9 & \textbf{61.3} & 53.2
& \textbf{29.5} & 29.3 & 29.1 \\
Qwen3-VL-8B-Instruct
& 78.4 & \textbf{81.8} & 70.4
& 70.1 & \textbf{70.9} & 70.7
& 55.3 & \textbf{55.9} & 53.9
& 78.0 & \textbf{78.5} & 72.9
& \textbf{34.8} & 33.3 & 33.7 \\
Qwen3-VL-4B-Instruct
& 68.1 & \textbf{72.8} & 61.5
& 66.5 & 66.8 & \textbf{67.4}
& \textbf{50.9} & 50.3 & 47.3
& \textbf{75.8} & 73.4 & 70.9
& 37.4 & \textbf{37.9} & 36.3 \\
Qwen3-VL-2B-Instruct
& 45.6 & \textbf{48.0} & 43.0
& 45.4 & \textbf{45.7} & 44.6
& 41.0 & 40.5 & \textbf{41.3}
& \textbf{54.1} & 52.3 & 47.6
& 36.5 & \textbf{36.9} & 36.3 \\
Qwen2.5-VL-7B-Instruct
& 64.8 & \textbf{69.7} & 58.2
& 56.9 & 56.5 & \textbf{58.8}
& 45.0 & \textbf{45.8} & 44.6
& \textbf{65.1} & 64.6 & 59.1
& 29.3 & 28.8 & \textbf{29.7} \\
Qwen2.5-VL-3B-Instruct
& 58.2 & \textbf{64.1} & 54.3
& 45.7 & 43.9 & \textbf{47.2}
& \textbf{43.1} & 42.4 & 41.3
& 56.2 & \textbf{58.2} & 53.6
& 28.0 & 28.5 & \textbf{28.7} \\
GLM-4.6V-Flash
& 85.5 & \textbf{90.9} & 80.1
& 69.9 & 68.1 & \textbf{70.5}
& \textbf{55.4} & 54.2 & 53.8
& \textbf{78.7} & 77.9 & 69.2
& \textbf{42.9} & 41.9 & 39.8 \\
Gemma3-12B-IT
& \textbf{27.1} & 25.5 & 26.6
& 58.6 & \textbf{60.4} & 59.5
& 35.4 & \textbf{35.6} & 35.0
& \textbf{30.8} & 30.5 & 30.1
& \textbf{31.9} & 31.6 & 30.1 \\
Gemma3-4B-IT
& 24.2 & 24.4 & \textbf{24.5}
& 34.5 & 34.3 & \textbf{35.2}
& \textbf{34.5} & 33.4 & \textbf{34.5}
& \textbf{26.3} & 25.7 & 25.5
& 28.4 & 29.0 & \textbf{29.2} \\
InternVL3.5-8B
& 81.0 & \textbf{87.5} & 79.3
& 64.3 & 64.6 & \textbf{64.8}
& \textbf{53.8} & 53.1 & 50.3
& 78.9 & \textbf{81.7} & 74.8
& 33.2 & \textbf{34.1} & 32.6 \\
InternVL3.5-4B
& 78.5 & \textbf{83.4} & 75.9
& 57.7 & \textbf{59.5} & 58.1
& \textbf{57.7} & 56.2 & 53.4
& 73.9 & \textbf{74.0} & 65.2
& 31.9 & 31.3 & \textbf{32.5} \\
InternVL3.5-2B
& 46.5 & \textbf{49.5} & 44.8
& 40.8 & \textbf{40.9} & 39.7
& 40.1 & \textbf{41.4} & 39.4
& 44.8 & \textbf{44.9} & 43.7
& 32.8 & 32.9 & \textbf{33.2} \\
\midrule
Kimi-K2.6
& 95.1 & \textbf{96.7} & 91.2
& 87.8 & \textbf{88.5} & 88.0
& 79.4 & \textbf{82.1} & 77.2
& 87.1 & \textbf{88.1} & 85.9
& \textbf{55.0} & 53.6 & 54.8 \\
Qwen3.6-plus
& 71.9 & \textbf{73.0} & 66.2
& 46.1 & \textbf{47.2} & 46.8
& 38.7 & \textbf{40.3} & 38.7
& \textbf{39.4} & 38.6 & 34.1
& 38.5 & \textbf{39.5} & 37.4 \\
Doubao-Seed-2.0-lite
& 83.0 & \textbf{86.0} & 77.0
& 49.0 & 49.0 & \textbf{51.0}
& 39.2 & 37.7 & \textbf{39.6}
& 42.8 & \textbf{43.8} & 42.3
& 33.8 & 34.1 & \textbf{35.2} \\
MiMo-V2.5
& 67.1 & \textbf{70.0} & 64.1
& 45.0 & 45.7 & \textbf{48.4}
& \textbf{48.5} & \textbf{48.5} & 48.4
& 55.6 & \textbf{55.9} & 51.3
& \textbf{37.1} & 36.6 & 35.1 \\
ERNIE 5.0
& 57.6 & \textbf{58.9} & 57.6
& \textbf{49.2} & 48.1 & 48.8
& 39.7 & 38.8 & \textbf{42.4}
& \textbf{36.4} & 35.9 & 35.2
& 33.6 & \textbf{33.8} & 32.3 \\
Gemini-3.5-Flash
& 90.8 & \textbf{92.8} & 82.9
& 80.5 & 80.7 & \textbf{81.0}
& \textbf{79.5} & 79.2 & 73.5
& \textbf{82.2} & 81.4 & 76.9
& \textbf{69.2} & 68.5 & 69.0 \\
GPT-5.5
& 73.8 & \textbf{75.7} & 68.9
& 66.8 & 67.7 & \textbf{68.9}
& \textbf{62.4} & 61.2 & 57.4
& 52.3 & \textbf{52.9} & 47.7
& \textbf{52.0} & 50.4 & 50.1 \\
\midrule
Avg.
& 68.2 & \textbf{71.2} & 64.2
& 58.7 & 58.9 & \textbf{59.4}
& \textbf{50.6} & 50.4 & 49.1
& \textbf{60.4} & 60.2 & 55.9
& \textbf{38.6} & 38.4 & 37.9 \\
\midrule
Human
& \textbf{100.0} & 98.3 & 99.2
& 95.8 & \textbf{97.5} & 93.3
& 96.7 & \textbf{98.3} & 93.3
& \textbf{95.0} & \textbf{95.0} & \textbf{95.0}
& 90.0 & \textbf{94.2} & 92.5 \\

\bottomrule
\end{tabular}
}
\end{table}

\subsection{Ruling Out Alternative Explanations}

The findings reported above are consistent with our central claim that stable sequential color ordering and sufficient lightness contrast constitute genuine, machine-readable signals for FM spatial reasoning. However, several alternative explanations could, in principle, account for the same pattern of results without implicating color design itself. Performance under degraded, non-idealized map images could differ from the clean, digitally rendered maps used throughout this study; the observed sensitivity to color encoding could instead reflect models' reliance on other task-solving strategies, such as legend-value matching or region localization, that happen to covary with color; and the degradation under randomized encoding could reflect a fixed ``darker-means-more'' prior rather than genuine ordinal reasoning over the color progression itself. We address each of these possibilities in turn below.
\subsubsection{Robustness to Map Degradation}
\label{sec:degradation}

All maps evaluated in the main experiments are clean, digitally rendered artifacts, whereas real-world thematic maps are frequently affected by geometric distortion, sensor noise, and resolution loss (e.g., scanned documents, photographed maps, or low-resolution reproductions). To assess whether the reported sensitivity to sequential color ordering generalizes beyond idealized rendering conditions, we conducted a degradation case study on InternVL3.5-8B, applying three single-factor input degradations to the sequential/randomized (H2) benchmark subset under the same dual-condition evaluation protocol used in the main experiments: (i) random rotation (31°--329°, reproducibly seeded per instance), (ii) additive Gaussian pixel noise ($\sigma=6$, approximately 2.4\% of the 0--255 range), and (iii) resolution reduction (1200$\times$1200 to 600$\times$600, corresponding to an effective DPI reduction from 220 to 110). Each degradation was applied in isolation to allow independent attribution of its effect.

Table~\ref{tab:degradation} summarizes the results. Gaussian noise had a negligible effect on overall accuracy (sequential: 59.1\%$\to$58.5\%; randomized: 46.3\%$\to$45.8\%) and preserved the sequential-randomized gap almost exactly (12.8 vs. 12.7 percentage points), indicating that the model's reliance on sequential color ordering is robust to mild pixel-level visual noise of the kind introduced by compression or scanning artifacts. In contrast, rotation and resolution reduction both caused substantial overall performance degradation (accuracy dropping by 21--23 percentage points under the sequential condition) and, notably, compressed the sequential-randomized gap considerably (from 12.8 points at baseline to 5.9 points under rotation and 4.2 points under resolution reduction). This suggests that once severe geometric distortion or resolution loss impairs the model's ability to parse spatial layout and region boundaries, performance degrades toward a common floor regardless of color encoding, and the additional benefit of preserved sequential ordering becomes proportionally smaller. Taken together, these results indicate that the sequential-ordering effect identified in this study is robust to the kind of low-level visual noise most representative of digitization artifacts, while more severe geometric or resolution degradation affects overall map legibility broadly enough to partially mask, rather than eliminate, the effect.

\begin{table}[H]
\centering
\small
\caption{InternVL3.5-8B accuracy (\%) under input degradation, compared with the clean-map baseline (Table~\ref{tab:sequential_effect}). $\Delta$ denotes the accuracy drop after disrupting sequential ordering.}
\label{tab:degradation}
\begin{tabular}{lccc}
\toprule
Condition & Sequential & Randomized & $\Delta$ \\
\midrule
Baseline (clean) & 59.1 & 46.3 & -12.8 \\
+ Gaussian noise & 58.5 & 45.8 & -12.7 \\
+ Rotation & 37.5 & 31.7 & -5.9 \\
+ Resolution reduction & 35.8 & 31.6 & -4.2 \\
\bottomrule
\end{tabular}
\end{table}

\subsubsection{Factorial Ablation of Attribute and Spatial Information}
\label{sec:matched-ablation}

The purpose of this experiment was to distinguish three potential sources of error in the benchmark: color-and-legend decoding, spatial-relation reasoning, and the integration of thematic and spatial information. We therefore conducted a matched $2 \times 2$ factorial experiment in which the availability of textual attribute information and explicit spatial information was manipulated independently. All four conditions used exactly the same 1,152 choropleth maps, questions, underlying values, and answer choices: (1) \emph{Full}, in which models recovered attribute values from the colors and legend and inferred spatial relations from the map; (2) \emph{Values supplied}, in which each region's numeric value and ordinal class were provided as a clean text list; (3) \emph{Spatial supplied}, in which region centroid coordinates and an adjacency list were provided; and (4) \emph{Both supplied}, in which both forms of auxiliary information were included. The spatial information was question-independent and did not state the answer directly, but provided explicit geometric and topological primitives from which task-relevant relations could be derived. Unlike the value-in-region control reported in Appendix~\ref{app:value-in-region}, the values were supplied as text rather than rendered inside the map, avoiding additional OCR and region--label association demands.

\begin{figure*}[t]
\centering
\includegraphics[width=\textwidth]{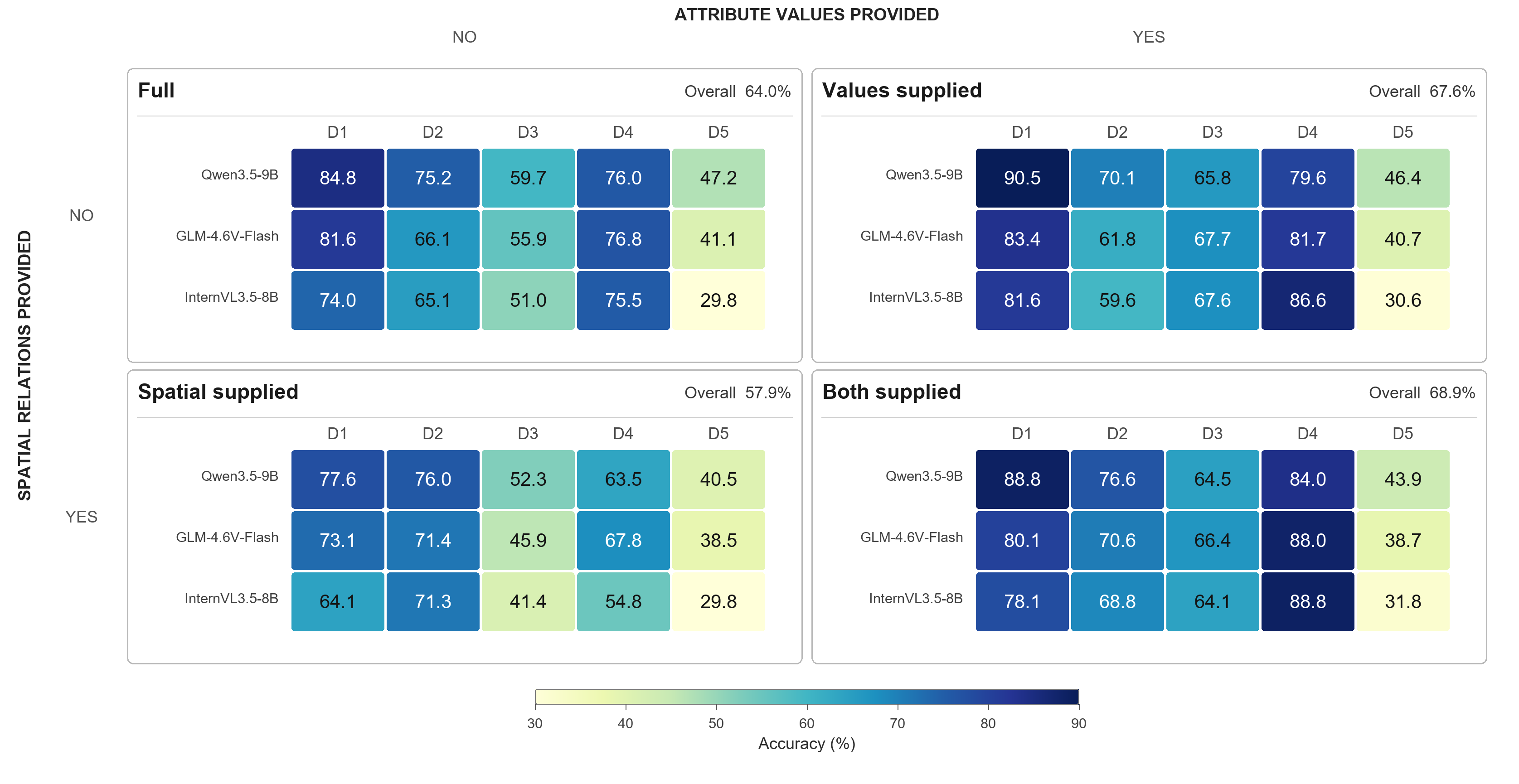}
\caption{Results of the matched $2 \times 2$ factorial experiment. Columns indicate whether region attribute values and ordinal classes were supplied as structured text, and rows indicate whether centroid coordinates and adjacency relations were supplied. Each quadrant reports accuracy (\%) for three models across the five task dimensions. Overall denotes accuracy averaged across the three models and five dimensions.}
\label{fig:matched-factorial}
\end{figure*}

We evaluated Qwen3.5-9B, GLM-4.6V-Flash, and InternVL3.5-8B under all four conditions across the five task dimensions. Figure.~\ref{fig:matched-factorial} reports the model- and dimension-level accuracies. The Both supplied condition achieved the highest overall accuracy for each model: Qwen3.5-9B increased from 68.6\% under Full to 71.6\%, GLM-4.6V-Flash from 64.3\% to 68.8\%, and InternVL3.5-8B from 59.1\% to 66.3\%. Averaged across the three models, overall accuracy was 64.0\% under Full, 57.9\% under Spatial supplied, 67.6\% under Values supplied, and 68.9\% under Both supplied.

The attribute-information contrasts show that color-and-legend decoding contributes measurably to the observed errors. Without spatial assistance, supplying clean textual attributes increased overall accuracy from 64.0\% to 67.6\% ($+3.6$ percentage points). When spatial information was held available, the corresponding contrast was considerably larger: Both supplied outperformed Spatial supplied by 11.0 points (68.9\% versus 57.9\%). At the dimension level, Values supplied improved D1 from 80.1\% to 85.2\%, D3 from 55.5\% to 67.0\%, and D4 from 76.1\% to 82.6\% relative to Full. Because these dimensions require identifying, comparing, or ranking thematic magnitudes, the improvements indicate that recovering attributes from colors and the legend is a non-trivial bottleneck. They do not imply that color is uninformative; rather, they quantify the benefit of bypassing perceptual decoding with directly accessible symbolic attributes.

The spatial-information contrasts depended on how the attributes were obtained. When attributes still had to be decoded from the map, adding centroid and adjacency information reduced overall accuracy from 64.0\% to 57.9\% ($-6.1$ points). Nevertheless, D2---the dimension most directly concerned with geometric and topological recognition---increased from 68.8\% to 72.9\%, confirming that the supplied representation contained useful spatial information. When attributes were instead supplied explicitly, adding the same spatial information increased overall accuracy from 67.6\% to 68.9\% ($+1.3$ points). It also improved D2 from 63.8\% to 72.0\% and D4 from 82.6\% to 87.0\%. Thus, spatial assistance was most useful when models did not simultaneously need to recover the associated attributes from map colors.

This reversal yields a positive interaction of approximately 7.4 percentage points:
$(68.9-57.9)-(67.6-64.0)=7.4$. The interaction indicates that errors cannot be attributed independently to color reading or spatial reasoning alone. Instead, a substantial part of the difficulty arises when models must align spatial information with attributes obtained through a different representational channel. The decline under Spatial supplied alone is therefore better interpreted as evidence of additional attention and cross-representational integration demands than as evidence that the supplied spatial information was irrelevant.

Finally, D5 remained difficult under all four conditions, with average accuracy ranging only from 36.3\% to 39.4\%. Supplying region-wise values, centroids, and local adjacency relations therefore did not resolve tasks requiring global spatial-structure delineation. Taken together, the experiment identifies color-and-legend decoding as one source of error, shows that explicit spatial primitives selectively assist directly spatial tasks, and reveals an interaction bottleneck when thematic and spatial information must be jointly integrated.
\subsubsection{Reversed Color Convention}
\label{sec:reverse-convention}
The performance decline under randomized encoding could in principle reflect reliance on a fixed ``darker-means-more'' prior rather than genuine ordinal reasoning over the presented color progression. To test this, we constructed a reversed-convention variant of the sequential benchmark in which lighter regions correspond to higher attribute values while preserving monotonic color progression, and re-evaluated InternVL3.5-8B, Qwen3.5-9B, and GLM-4.6V-Flash. Table~\ref{tab:reverse_convention} reports overall accuracy.

\begin{figure}[H]
\centering
\includegraphics[width=\textwidth]{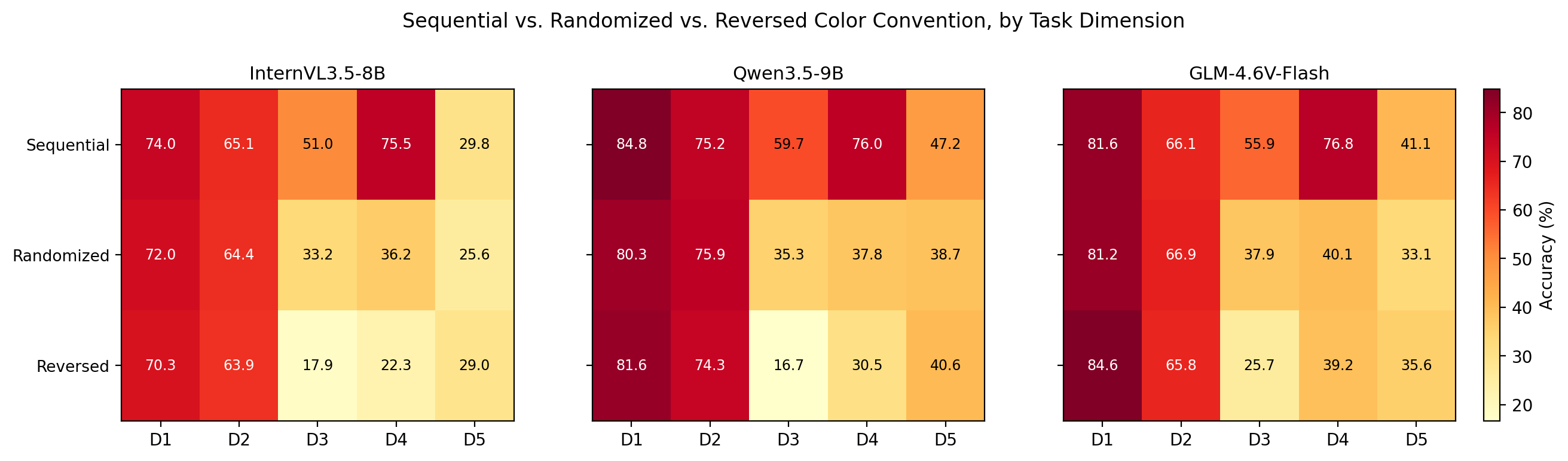}
\caption{Accuracy (\%) under original, randomized, and reversed color conventions, by task dimension (D1--D5), for InternVL3.5-8B, Qwen3.5-9B, and GLM-4.6V-Flash.}
\label{fig:reversed_by_dim}
\end{figure}

Overall, accuracy under the reversed convention is lower than under randomized encoding for all three models, suggesting that models are not simply indifferent to color direction. The dimension-level results in Fig.~\ref{fig:reversed_by_dim} reveal why, with a highly consistent pattern across all three models. For D1 (attribute identify) and D2 (spatial recognition), accuracy under the original, randomized, and reversed conditions is nearly indistinguishable, consistent with these dimensions' established independence from color encoding. For D3 (compare) and D4 (rank), however, all three models show a consistent ordering of Original $>$ Randomized $>$ Reversed, with the reversed convention producing the lowest accuracy of all three conditions (e.g., D3: 51.0/33.2/17.9\% for InternVL3.5-8B; 59.7/35.3/16.7\% for Qwen3.5-9B; 55.9/37.9/25.7\% for GLM-4.6V-Flash, respectively). This indicates that for tasks requiring explicit magnitude comparison, models have partially internalized a directional ''darker-means-more'' prior, and a consistently inverted convention actively misleads this prior, producing more systematic errors than the unstructured degradation caused by randomization.

\begin{table}[H]
\centering
\small
\caption{Overall accuracy (\%) under the original sequential and reversed (lighter = higher) sequential color conventions, compared with randomized encoding.}
\label{tab:reverse_convention}
\begin{tabular}{lccc}
\toprule
Model & Sequential & Randomized & Reversed \\
\midrule
InternVL3.5-8B & 59.1 & 46.3 & 40.7 \\
Qwen3.5-9B & 68.6 & 53.6 & 48.7 \\
GLM-4.6V-Flash & 64.3 & 51.9 & 50.2 \\
\bottomrule
\end{tabular}
\end{table}

Taken together, these results indicate that FM sensitivity to sequential color ordering is not attributable to a single mechanism. For tasks that are largely independent of color encoding, the direction of the color-value mapping produces comparatively small changes in accuracy relative to D3 and D4. In contrast, for tasks requiring explicit magnitude comparison and ranking, models rely on a direction-specific ``darker-means-more'' prior, for which reversing the convention is substantially more disruptive than removing ordinal structure altogether. This dimension-dependent pattern is highly consistent across all three evaluated models, suggesting it reflects a general property of how FMs process sequential choropleth color encoding rather than an idiosyncrasy of a single architecture.

\subsection{Are Cartographic Sensitivities Intrinsic or Learnable?}

\begin{figure} [H]
	\centering
	\includegraphics[width=\textwidth]{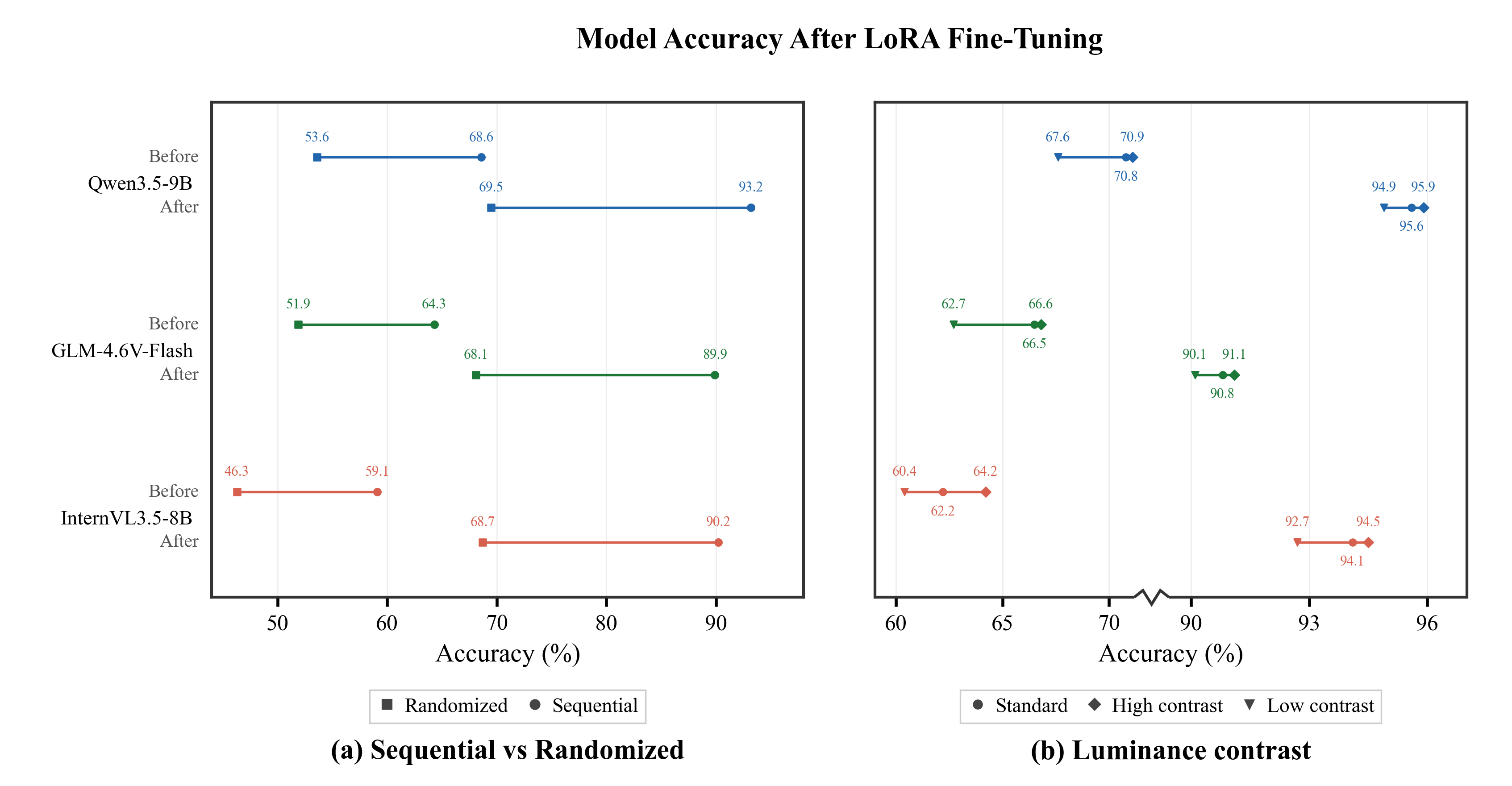}
        \captionsetup{skip=0pt}  % 调整该图的上下间距
	\caption{Accuracy before and after LoRA fine-tuning under different 
         cartographic conditions for QWEN3.5-9B, GLM-4.6V-Flash, and InternVL3.5-8B. (a) Sequential vs. randomized. (b) Standard, high-contrast, and low-contrast luminance conditions.}
	\label{lora}
\end{figure}

After establishing that sequential ordering and luminance contrast significantly influence FM geospatial reasoning, we further investigate whether these cartographic sensitivities are intrinsic properties of FMs or can instead be learned through task-specific adaptation. To explore this question, we conduct LoRA fine-tuning experiments on the models that achieved the best overall performance within their respective model families in Table~\ref{tab:hue_effect}, \ref{tab:sequential_effect}, \ref{tab:luminance_effect}, including Qwen3.5-9B, GLM-4.6V-Flash, and InternVL3.5-8B \cite{shuttleworth2026lora}. To avoid introducing artificial bias toward any particular cartographic design, the training corpus maintained the same distribution of cartographic conditions as the benchmark, but using a completely disjoint set of thematic data and map instances. The complete training corpus contained approximately twice as many samples as the evaluation benchmark (4608 maps and 23040 QA pairs for H2 training, 6912 maps and 34560 QA pairs for H3 training). The results before and after finetune are shown in Figure~\ref{lora}.

The results reveal an important phenomenon. LoRA fine-tuning substantially improves overall geospatial reasoning performance across all evaluated models and cartographic conditions. Under sequential/randomized encoding settings, all fine-tuned models exhibit large performance gains compared with their original counterparts. For example, Qwen3.5-9B improves from 68.6\% to 93.2\% under sequential encoding, while its randomized performance also increases from 53.6\% to 69.5\%. Similar improvements are observed for GLM-4.6V-Flash and InternVL3.5-8B. Likewise, under luminance contrast conditions, all models exhibit substantial performance improvement after adaptation. For example, Qwen3.5-9B increases from 70.8\% to 95.6\% under the standard contrast condition, while GLM-4.6V-Flash improves from 66.5\% to 90.8\%. These results indicate that a large portion of choropleth reasoning capability can indeed be acquired efficiently through lightweight task-specific adaptation.

However, despite these substantial performance improvements, cartographic sensitivities do not disappear after fine-tuning. Under sequential/randomized conditions, all evaluated models continue to exhibit clear degradation after sequential ordering is disrupted. For instance, Qwen3.5-9B still shows a 13.7-point degradation after LoRA fine-tuning, while GLM-4.6V-Flash and InternVL3.5-8B continue to exhibit drops of 11.8 and 21.5 points, respectively. Similarly, under luminance contrast manipulation, low-contrast settings consistently remain weaker than standard- or high-contrast conditions even after adaptation. Although the magnitude of degradation becomes smaller after fine-tuning, the overall sensitivity pattern remains largely unchanged. These findings suggest that cartographic sensitivities are neither purely intrinsic nor entirely learnable. On the one hand, FMs can rapidly acquire substantially improved choropleth reasoning capability through lightweight adaptation, indicating that many map interpretation strategies are learnable. On the other hand, the persistent influence of sequential ordering and luminance separability after adaptation suggests that these cartographic principles continue to function as relatively stable structural priors during machine spatial reasoning. More broadly, the results imply that AI-friendly cartographic design may remain important even for future task-adapted geospatial FMs, since certain visual organizations appear fundamentally easier for machines to interpret.

\subsection{Implications for AI-Friendly Cartography}

The results of this study suggest that not all classical cartographic design principles contribute equally to FM spatial reasoning. More importantly, the findings reveal that some cartographic principles originally developed for human map reading also remain highly beneficial for machine-based spatial understanding. These observations provide several important implications for the emerging direction of AI-friendly cartography.

\textbf{First, AI-friendly thematic maps should preserve conventional sequential visual organization.} The experiments consistently show that disrupting sequential ordering substantially degrades model reasoning performance, particularly for comparison-based thematic reasoning tasks. Although we originally hypothesized that FMs might primarily rely on isolated local visual tokens, the large performance degradation caused by randomized encoding indicates that models also depend heavily on consistent thematic ordering. The reversed-palette results further suggest that this benefit depends on the familiar forward mapping (darker means higher), not on ordinal structure alone. Sequential choropleth maps provide stable visual correspondences between lightness progression and thematic magnitude, enabling models to more effectively infer regional hierarchies, thematic transitions, and relative spatial relationships. These findings suggest that FMs are not merely performing low-level pixel discrimination but are also capable of utilizing higher-level visual organization embedded in cartographic representations.

\textbf{Second, AI-friendly cartography should maintain sufficient lightness separability between neighboring thematic classes and, when accurate attribute identification is particularly important, may further enlarge lightness differences intentionally.} Compared with hue variation, manipulating lightness contrast produces substantially stronger and more systematic performance variation across both models and task dimensions. In particular, reducing lightness contrast consistently degrades model reasoning performance across most evaluated models, especially for local thematic identification tasks. This finding indicates that maintaining sufficient regional separability is critical for reducing thematic ambiguity and supporting reliable machine spatial reasoning. Moreover, although increasing contrast beyond the standard setting only produces relatively limited overall improvement, the results suggest that further enlarging lightness differences can still enhance attribute identification accuracy in certain tasks and models. Therefore, AI-oriented thematic map design should not only avoid insufficient luminance contrast, but may also intentionally employ stronger lightness separability when accurate thematic attribute recognition is particularly important.

\textbf{Third, AI-friendly thematic maps may place less emphasis on specific hue semantics and aesthetic palette variation.} Traditional thematic cartography often emphasizes hue semantics, perceptual harmony, and aesthetic consistency. However, our experiments show that changing sequential hue palettes produces only limited performance variation across FMs. This finding suggests that FMs rely more heavily on structural visual organization than on semantic color associations commonly used in human map interpretation. Consequently, some color design principles primarily targeting human aesthetics may transfer less effectively to machine cognition.

More broadly, the results indicate that human-centered cartography and machine-centered cartography are neither fully identical nor completely independent. Certain cartographic principles, such as sequential ordering and sufficient lightness separability, appear to simultaneously facilitate both human and machine spatial reasoning. In contrast, other principles that mainly target human aesthetics or semantic color associations may transfer less effectively to machine cognition. These observations suggest that future cartography research may need to move beyond purely human-centered design assumptions and begin systematically investigating which visual encoding principles are fundamentally machine-readable.

\subsection{Limitations and Future Work}

Although this study provides a systematic investigation of how choropleth color encoding influences FM spatial reasoning, several limitations remain and deserve further exploration.

First, the current study focuses primarily on sequential choropleth maps. While sequential encoding constitutes one of the most commonly used forms of thematic cartography, many real-world geographic visualizations also employ diverging, qualitative, bivariate, and uncertainty-aware color schemes. Different cartographic representations may involve substantially different visual organization principles and cognitive mechanisms for both humans and machines. Future work should therefore investigate whether the findings observed in this study generalize to broader categories of thematic map design.

Second, the present benchmark mainly evaluates model performance through final-answer accuracy. Although the experimental results reveal clear sensitivity patterns to cartographic color organization, they do not directly explain the internal reasoning mechanisms of FMs during map understanding. Future work could combine attention analysis, token attribution, visual saliency analysis, and mechanistic interpretability methods to further investigate how models internally process cartographic visual structures and thematic encoding patterns.

Third, the current experiments focus on static map understanding under controlled benchmark settings. However, many real-world cartographic applications involve interactive visualization, dynamic map exploration, and multi-scale spatial navigation. Future research could therefore investigate how FMs interact with dynamic and interactive cartographic environments, particularly in agent-based geographic reasoning systems and embodied spatial cognition tasks.

Finally, this work mainly focuses on color organization in choropleth maps, while many other cartographic variables may also influence machine spatial reasoning, including symbolization, legend design, label placement, visual hierarchy, map layout, and annotation structure. Future research could therefore expand beyond color design and systematically investigate broader principles of AI-friendly cartography, aiming to establish a more comprehensive theoretical framework for machine-oriented cartographic representation.

\section{Conclusion}

This study systematically investigated how classical choropleth color design principles influence FM spatial reasoning. Focusing on three representative cartographic factors---sequential hue palettes, sequential ordering, and lightness contrast---we constructed a large-scale controlled benchmark containing 5,760 choropleth maps and 28,800 spatial reasoning tasks, and evaluated 21 recent multimodal FMs under unified experimental settings. The results reveal that different cartographic color principles contribute unequally to machine spatial reasoning. First, sequential hue variation produces only a limited and non-systematic influence on model performance, suggesting that FMs rely less on hue semantics than human map readers. Second, disrupting sequential color ordering substantially degrades reasoning performance, particularly for comparison- and ranking-related tasks, indicating that conventional sequential organization serves as an important visual prior for machine reasoning. Third, lightness contrast constitutes a fundamental machine-readable signal. Reducing contrast consistently harms reasoning performance across most evaluated models, whereas further increasing contrast provides only limited additional benefit once sufficient regional separability is achieved. More broadly, the findings suggest that some classical cartographic principles originally developed for human perception remain highly beneficial for machine cognition, while others transfer less effectively to FM reasoning. In particular, FMs appear to depend more heavily on conventional sequential organization and regional separability than on semantic hue interpretation. These observations highlight the potential importance of AI-friendly cartography, where maps are designed not only for human readability but also for machine interpretability. More importantly, they suggest that improving machine spatial reasoning may depend not only on advancing models themselves, but also on optimizing the visual representations they interpret.

However, several limitations still remain in the current study. The benchmark mainly focuses on sequential choropleth maps and color-related variables, while other map types and cartographic elements remain unexplored. In addition, the current evaluation primarily targets general-purpose multimodal FMs rather than map-specialized systems. Future work can further investigate broader cartographic variables, additional thematic map types, and machine-oriented map design principles for AI spatial reasoning systems.

\section*{Acknowledgement(s)}

\section*{Disclosure statement}

No potential conflict of interest was reported by the author(s).

\section*{Data availability statement}

The datasets and code used in this study are available at GitHub: https://github.com/Myantion/CHROMA. The data and code are freely accessible and can be used for the purpose of reproducing the results (CC BY 4.0).

\section*{Funding}

This work was supported by grants from the National Natural Science Foundation of China (No. 42501551, 42371455) and the Tobii China Innovation Initiative Project (TPI250407CN).

\section*{Notes on contributor(s)}

\textbf{Yonghe Sun}: Methodology, Data curation, Software, Validation, Visualization, Writing - Original draft preparation. 
\textbf{Zhenjia Liu}: Methodology, Data curation, Software, Visualization, Writing – review editing. 
\textbf{Hua Liao}: Writing - Review \& Editing, Supervision, Project administration, Funding Acquisition.
\textbf{Wenjia Xu}: Writing - Review \& Editing, Supervision. 
\textbf{Nai Yang}: Writing - Review \& Editing, Supervision.
\textbf{Weihua Dong}: Writing - Review \& Editing, Supervision.  
\textbf{Zhiwei Wei}: Conceptualization, Methodology, Investigation, Writing - Original draft preparation, Supervision, Project administration, Funding Acquisition. 

\bibliographystyle{unsrtnat}
\bibliography{interacttfqsample}

\clearpage

\appendix
\section{Control Baselines for Template and Answer-Position Bias}
\label{app:baselines}

To verify that model performance in the main experiments reflects genuine map-reading capability rather than template or answer-position shortcuts, we conducted three control baselines on InternVL3.5-8B and Qwen3.5-9B, covering all five task dimensions (D1--D5): (i) a no-image condition, in which models received only the question text and answer options without any map image; (ii) a blank-image condition, in which a blank white image replaced the map; and (iii) a shuffled-answer condition combined with the blank image, in which answer options were additionally randomly reordered. As shown in Table~\ref{tab:app_baseline}, accuracy under all three conditions remained close to the chance level expected under random guessing. Four of the five task dimensions offer four answer options (25\% chance level), while D3 (attribute comparison) offers three options and part of D5 (Delineate) offers two options; with each dimension contributing one question per map, the overall weighted chance level is $(4512\times25\% + 1200\times33.3\% + 288\times50\%)/6000 \approx 27.8\%$, closely matching the observed baseline accuracy of approximately 27\% across models and conditions. This confirms that models cannot solve the benchmark questions through template recognition or answer-position shortcuts, and that the substantially higher accuracy observed in the main experiments (Secs.~4.1--4.3) genuinely reflects the use of map content.

\begin{table}[H]
\centering
\small
\caption{Accuracy (\%) under no-image, blank-image, and shuffled-answer control baselines, averaged across all five task dimensions (D1--D5). Weighted chance level $\approx$ 27.8\%.}
\label{tab:app_baseline}
\begin{tabular}{lccc}
\toprule
Model & No-Image & Blank-Image & Shuffled + Blank-Image \\
\midrule
InternVL3.5-8B & 28.1 & 27.2 & 27.1 \\
Qwen3.5-9B & 27.8 & 27.5 & 27.2 \\
\midrule
Avg. & 27.95 & 27.35 & 27.15 \\
\bottomrule
\end{tabular}
\end{table}

\section{Accuracy by Individual Question Subtype}
\label{app:subtype}

\begin{figure}[H]
\centering
\includegraphics[width=\textwidth]{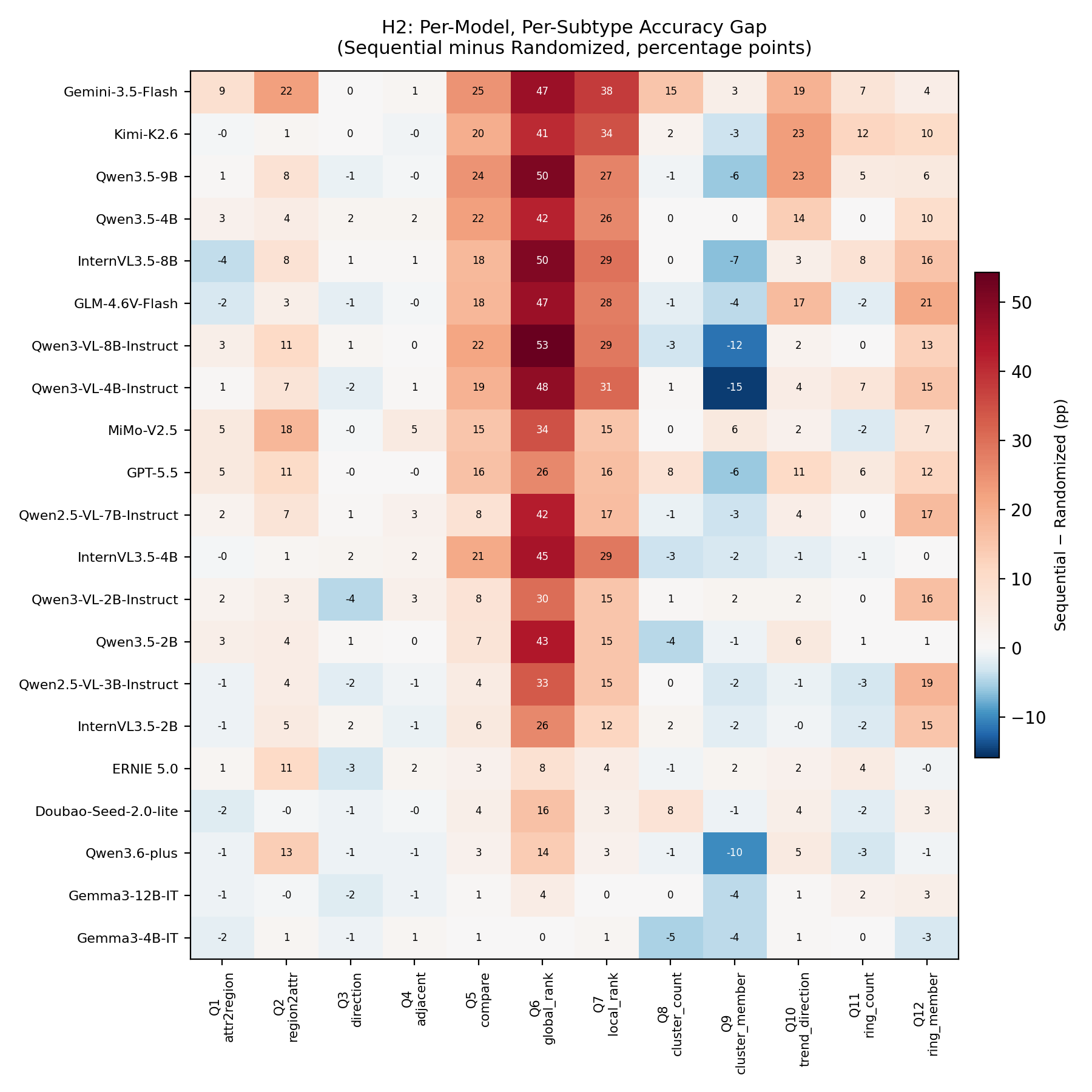}
\caption{Per-model, per-subtype accuracy gap between sequential and randomized encoding (percentage points; positive = higher accuracy under sequential encoding). Models are ordered by mean gap (largest at top). Subtype labels follow Table~\ref{tab:task-details}.}
\label{fig:h2_subtype_heatmap}
\end{figure}

\begin{figure}[H]
\centering
\includegraphics[width=\textwidth]{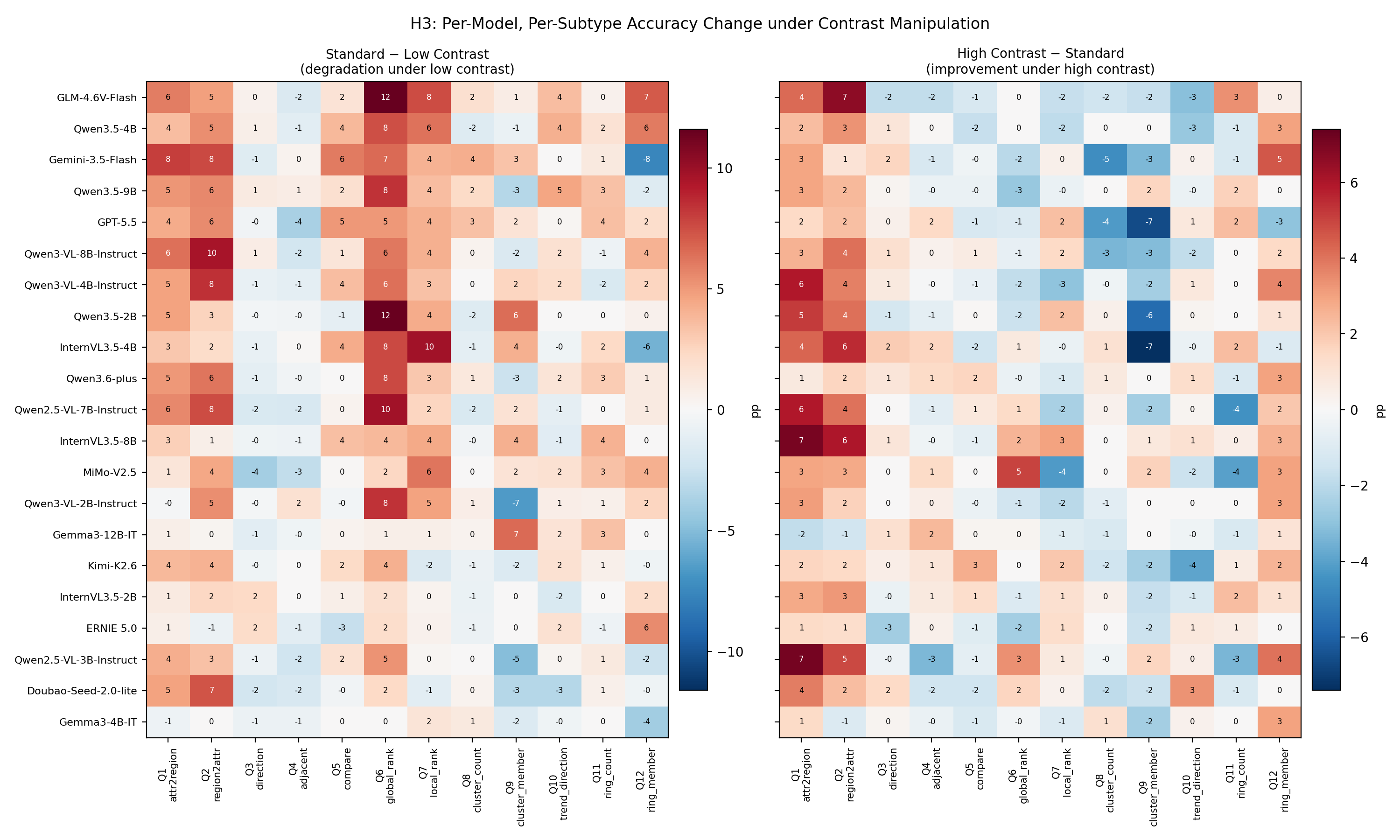}
\caption{Per-model, per-subtype accuracy change under contrast manipulation. Left: Standard minus Low Contrast (degradation under low contrast). Right: High Contrast minus Standard (improvement under high contrast). Models are ordered by mean Standard-minus-Low gap (largest at top).}
\label{fig:h3_subtype_heatmap}
\end{figure}

To complement the five-dimension aggregation (D1--D5) reported in the main text, Figures~\ref{fig:h2_subtype_heatmap} and~\ref{fig:h3_subtype_heatmap} report the full per-model, per-subtype (Q1--Q12) results for H2 and H3, respectively, visualized as heatmaps for readability.

Figure~\ref{fig:h2_subtype_heatmap} shows that the degradation under randomized encoding is highly concentrated in Q6 (global rank) and Q7 (local rank), and to a lesser extent Q5 (attribute comparison) — consistent with the D3/D4 sensitivity reported in Sec.~5.1 — while Q3 (direction) and Q4 (adjacent), corresponding to D2, show negligible change across nearly all models. Figure~\ref{fig:h3_subtype_heatmap} shows that both the degradation under low contrast (left panel) and the improvement under high contrast (right panel) are concentrated in Q1 and Q2 (attribute identification, D1) and, for low contrast, additionally in Q6, consistent with the D1 sensitivity reported in Sec.~5.2.

\section{Example Maps: Source Data and Benchmark Rendering}
\label{app:example_maps}

To further illustrate the benchmark construction, this appendix presents example maps at two stages: the real-world source thematic data prior to synthetic rendering, and the resulting benchmark maps across the hue, ordering, and contrast manipulations.

\subsection{Example Source Thematic Data}
\begin{figure}[H]
\centering
\includegraphics[width=\textwidth]{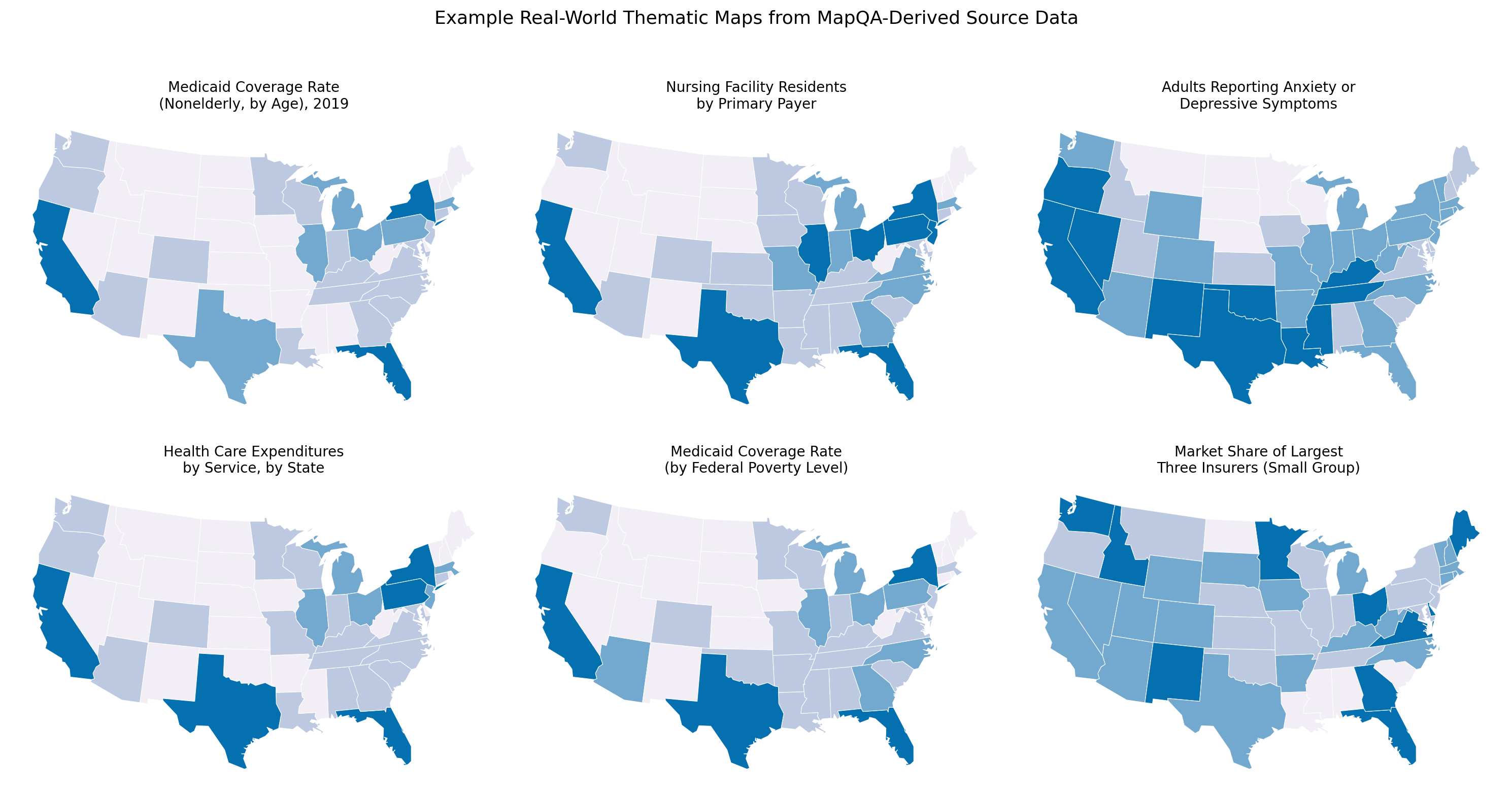}
\caption{Example real-world thematic maps from the MapQA-derived source data (Kaiser Family Foundation health and healthcare indicators), rendered at the US state level prior to the controlled synthetic map generation pipeline.}
\label{fig:source_thematic}
\end{figure}

Figure~\ref{fig:source_thematic} presents example real-world thematic maps from the MapQA-derived source data~\cite{chang2022mapqa} used to assign attribute values during benchmark construction, prior to the controlled synthetic rendering pipeline. These examples illustrate the diversity of real-world spatial distributions (e.g., health insurance coverage, healthcare expenditure, and mental health indicators across US states) from which the underlying thematic value structures were derived.

\subsection{Example Benchmark Map Renderings}

\begin{figure}[H]
\centering
\includegraphics[width=\textwidth]{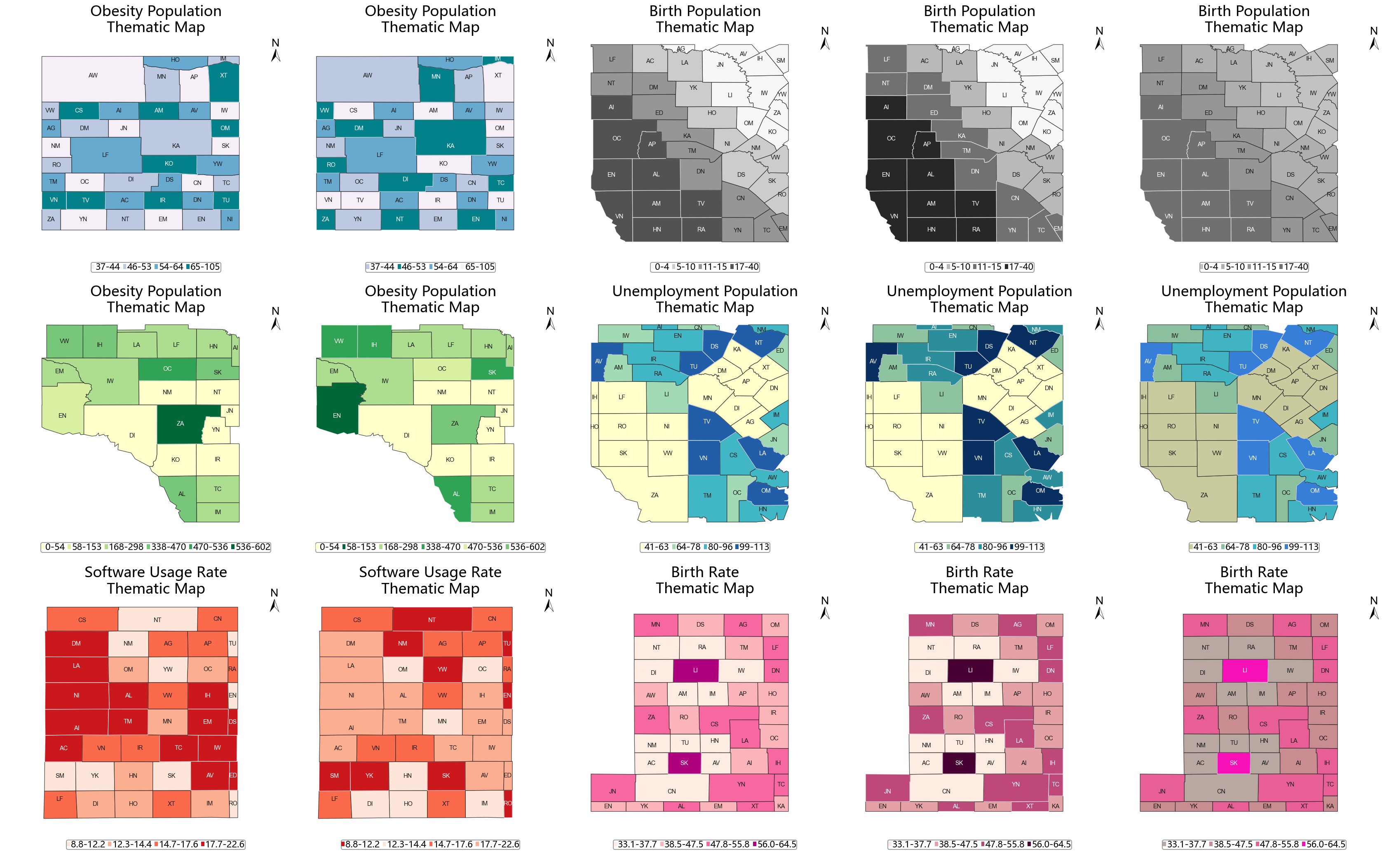}
\caption{Additional example choropleth maps from CHROMA, spanning five thematic categories (Obesity Population, Software Usage Rate, Birth Population, Unemployment Population, Birth Rate) and multiple color rendering variants used in the benchmark construction pipeline.}
\label{fig:example_maps_grid}
\end{figure}

Figure~\ref{fig:example_maps_grid} presents additional example maps from the CHROMA benchmark, illustrating the diversity of thematic content (e.g., obesity prevalence, software usage rate, birth rate, unemployment rate) and color rendering variants used across the benchmark's controlled color manipulations.

\section{Value-in-Region Experiment}
\label{app:value-in-region}

As a supplementary control, we constructed a color-free variant of the benchmark in which numerical attribute values were written directly inside their corresponding regions. We evaluated InternVL3.5-8B, Qwen3.5-9B, and GLM-4.6V-Flash on this value-in-region representation across all five task dimensions (D1--D5), using the same underlying maps, attribute values, questions, and answer choices as in the sequential color-encoding condition.

\begin{table}[H]
\centering
\scriptsize
\caption{Supplementary comparison of accuracy (\%) under the sequential color-encoding condition (Color) and the value-in-region, color-free condition (Value) across the five task dimensions (D1--D5). The value-in-region condition additionally requires numerical-text recognition and region--value association.}
\label{tab:app_color_value}
\resizebox{\textwidth}{!}{
\begin{tabular}{lcccccccccc}
\toprule
& \multicolumn{2}{c}{D1}
& \multicolumn{2}{c}{D2}
& \multicolumn{2}{c}{D3}
& \multicolumn{2}{c}{D4}
& \multicolumn{2}{c}{D5} \\
\cmidrule(lr){2-3}
\cmidrule(lr){4-5}
\cmidrule(lr){6-7}
\cmidrule(lr){8-9}
\cmidrule(lr){10-11}
Model
& Color & Value
& Color & Value
& Color & Value
& Color & Value
& Color & Value \\
\midrule
Qwen3.5-9B
& \textbf{84.8} & 64.0
& 75.2 & \textbf{76.2}
& \textbf{59.7} & 52.4
& \textbf{76.0} & 55.4
& \textbf{47.2} & 38.5 \\
GLM-4.6V-Flash
& \textbf{81.6} & 54.6
& 66.1 & \textbf{67.2}
& \textbf{55.9} & 53.8
& \textbf{76.8} & 62.5
& \textbf{41.1} & 36.5 \\
InternVL3.5-8B
& \textbf{74.0} & 49.1
& \textbf{65.1} & 63.4
& \textbf{51.0} & 53.0
& \textbf{75.5} & 61.3
& \textbf{29.8} & 27.5 \\
\midrule
Avg.
& \textbf{80.1} & 55.9
& 68.8 & \textbf{68.9}
& \textbf{55.5} & 53.1
& \textbf{76.1} & 59.7
& \textbf{39.4} & 34.2 \\
\bottomrule
\end{tabular}
}
\end{table}

As shown in Table~\ref{tab:app_color_value}, the value-in-region condition produces lower accuracy than the color-encoding condition on most task dimensions. The largest difference occurs for D1, where average accuracy decreases from 80.1\% to 55.9\% (-24.2 percentage points). D4 also exhibits a substantial decrease from 76.1\% to 59.7\% (-16.4 points), while D5 decreases from 39.4\% to 34.2\% (-5.2 points). In comparison, D2 shows virtually no difference (68.8\% versus 68.9\%), and D3 exhibits only a modest difference (55.5\% versus 53.1\%).

These results indicate that replacing color patches with numerical labels does not necessarily simplify the task for FMs. For D1 and D4, the color gradient may provide a perceptually salient cue for identifying attribute classes or extreme values, whereas the value-in-region representation requires models to recognize and compare multiple numerical labels distributed across the map. Similarly, color patches may make spatial clusters and structural patterns more visually salient for D5. Nevertheless, because the value-in-region condition introduces OCR and region--value association demands, these results should not be interpreted as a clean causal estimate of the benefit of color encoding. The matched four-condition experiment reported in Sec.~\ref{sec:matched-ablation} addresses these confounds by supplying attribute values and spatial information as clean auxiliary text while keeping the map images, questions, and answer choices unchanged.

\end{document}